%% file: arxiv.tex
\documentclass{article}

\usepackage[preprint]{neurips_2026}
\usepackage{amsmath}
\usepackage{multirow}
\usepackage{adjustbox}
\usepackage{wrapfig}

\newcommand{\dnaemoji}{%
  \raisebox{-0.15em}{\includegraphics[height=1em]{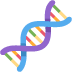}}%
}

\usepackage[utf8]{inputenc} % allow utf-8 input
\usepackage[T1]{fontenc}    % use 8-bit T1 fonts
\usepackage{hyperref}       % hyperlinks
\usepackage{url}            % simple URL typesetting
\usepackage{booktabs}       % professional-quality tables
\usepackage{amsfonts}       % blackboard math symbols
\usepackage{nicefrac}       % compact symbols for 1/2, etc.
\usepackage{microtype}      % microtypography
\usepackage{xcolor}         % colors
\usepackage{graphicx}
\usepackage{amssymb}
\usepackage{amsthm}

\usepackage{listings}
\usepackage{xcolor}

\lstdefinestyle{promptstyle}{
  basicstyle=\ttfamily\small,
  breaklines=true,
  breakatwhitespace=false,
  columns=fullflexible,
  keepspaces=true,
  frame=single,
  xleftmargin=1em,
  xrightmargin=1em,
  aboveskip=1em,
  belowskip=1em
}

\NewDocumentCommand{\gabri}
{ mO{} }{\textcolor{red}{\textsuperscript{\textit{Gabri}}\textsf{\textbf{\small[#1]}}}}

\NewDocumentCommand{\edward}
{ mO{} }{\textcolor{green}{\textsuperscript{\textit{Edward}}\textsf{\textbf{\small[#1]}}}}

\NewDocumentCommand{\carl}
{ mO{} }{\textcolor{blue}{\textsuperscript{\textit{Carl}}\textsf{\textbf{\small[#1]}}}}

\newcommand{\at}[1]{@\texorpdfstring{$#1$}{#1}}

\NewDocumentCommand{\alex}
{ mO{} }{\textcolor{orange}{\textsuperscript{\textit{Alex}}\textsf{\textbf{\small[#1]}}}}

\title{AssayBench: An Assay-Level Virtual Cell Benchmark for LLMs and Agents}%ScreensQA - Biology's Last Exam}

\author{%
  Edward De Brouwer\dnaemoji \And Carl Edwards\dnaemoji \And Alexander Wu\dnaemoji \AND Jenna Collier \And Graham Heimberg \And Xiner Li \AND Meena Subramaniam \And Ehsan Hajiramezanali \And David Richmond \AND Jan-Christian H{\"u}tter \And Sara Mostafavi \And Gabriele Scalia \AND
  Genentech\\
  South San Francisco, CA, USA\\
    \texttt{\{debroue1,edwarc24,wua33,scaliag\}@gene.com} \\
  \dnaemoji \small  These authors contributed equally.
}

\usepackage{xspace}

\newcommand{\screensqa}{\textsc{AssayBench}\xspace}

\newcommand{\GeminiPro}{\textsc{Gemini 3 Pro}\xspace}

\newcommand{\novel}{\texttt{LaTest}\xspace}

\newcommand{\andcg}{\textrm{AnDCG}}
\newcommand{\ndcg}{\textrm{nDCG}}
\newcommand{\dcg}{\textrm{DCG}}

\newcommand{\Prec}{\textrm{Precision}}
\newcommand{\fdr}{\textrm{dFDR}}

\begin{document}

\maketitle

\input{sections/0abstract}

%\carl{Example of a comment}. \edward{}. \gabri{}. \alex{}.

\input{sections/1introduction}
\input{sections/2data}
\input{sections/3metrics}

\input{sections/4results}

\input{sections/5discussion}

\clearpage

\bibliographystyle{plainnat} \bibliography{references} 

%%%%%%%%%%%%%%%%%%%%%%%%%%%%%%%%%%%%%%%%%%%%%%%%%%%%%%%%%%%%
\clearpage
\appendix

\input{sections/appendix/A_dataprep}

\input{sections/appendix/B_metrics}

\input{sections/appendix/C_benchmarks}

\input{sections/appendix/D_results}

\end{document}

%% file: sections/0abstract.tex
\begin{abstract}

Recent advances in machine learning and large-scale biological data collections have revived the prospect of building a virtual cell, a computational model of cellular behavior that could accelerate biological discovery. One of the most compelling promises of this vision is the ability to perform in silico phenotypic screens, in which a model predicts the effects of cellular perturbations in unseen biological contexts. This task combines heterogeneous textual inputs with diverse phenotypic outputs, making it particularly well-suited to LLMs and agentic systems. 
Yet, no standard benchmark currently exists for this task, as existing efforts focus on narrower molecular readouts that are only indirectly aligned with the phenotypic endpoints driving many real-world drug discovery workflows. In this work, we present \screensqa{}, a benchmark for phenotypic screen prediction, built from 1{,}920 publicly available CRISPR screens spanning five broad classes of cellular phenotypes. We formulate the screen prediction task as a gene rank prediction for each screen and introduce the adjusted \ndcg{}, a continuous metric for comparing performance across heterogeneous assays. Our extensive evaluation shows that existing methods remain far from empirically estimated performance ceilings and zero-shot generalist LLMs outperform biology-specific LLMs and trainable baselines. Optimization techniques such as fine-tuning, ensembling, and prompt optimization can further improve LLM performance on this task. Overall, \screensqa{} offers a practical testbed for measuring progress toward in-silico phenotypic screening and, more broadly, virtual cell models. Our benchmark is available at~\url{https://github.com/Genentech/AssayBench}.

\end{abstract}

%% file: sections/1introduction.tex
\section{Introduction}

%Recent progress in machine learning and large-scale biological data generation has revived interest in the virtual cell: a computational model that captures cellular responses across perturbations, contexts, and readouts \cite{bunne2024build, roohani2025virtual}. If realized, such models could transform early-stage drug discovery by enabling researchers to anticipate the consequences of genetic or chemical interventions before running experiments. Among the most compelling applications of this vision is the ability to perform in silico phenotypic screens, where a model predicts which perturbations are likely to modulate a cellular phenotype of interest. In contrast to purely molecular prediction tasks like perturbSeq, phenotypic screening is directly tied to many practical discovery workflows, where the goal is not only to characterize cellular state but to identify actionable perturbations that produce a desired functional outcome~\citep{vincent2022phenotypic,replogle2022mapping}.

Recent progress in machine learning and large-scale biological data generation has renewed interest in the \emph{virtual cell}: a computational model that predicts how cells respond across perturbations, contexts, and readouts~\citep{bunne2024build,roohani2025virtual}. If realized, such models would substantially accelerate early drug discovery by enabling researchers to anticipate the effects of genetic or chemical interventions before running experiments. One of the most compelling applications of this vision is \emph{in silico phenotypic screening}, in which a model predicts which perturbations are most likely to modulate a phenotype of interest. Unlike tasks that focus on reconstructing the molecular states of cells, phenotypic screening is directly aligned with the decision problems encountered in the drug discovery pipeline~\citep{vincent2022phenotypic,replogle2022mapping}. For example, identifying which genetic perturbations decrease resistance to etoposide, a chemotherapeutic agent, in KBM-7 chronic myeloid leukemia cells could inform the development of combination therapies to overcome drug resistance in leukemia patients~\citep{wang2014genetic}.

%Despite this promise, progress toward in silico phenotypic screening remains difficult to evaluate systematically. Existing benchmarks for biological perturbation modeling have largely focused on molecular readouts, especially transcriptomic responses measured with single-cell assays~\citep{peidli2024scperturb,wu2024perturbench,wu2025contextualizing,youngblut2025scbasecount}. These resources have been invaluable for studying perturbation effects at scale, but they only partially capture the downstream screening settings that matter in drug discovery. In a phenotypic screen, success is typically defined at the level of an assay endpoint: survival, proliferation, infection burden, reporter activity, trafficking, or another functional phenotype. The relevant prediction problem is therefore not simply to reconstruct molecular state, but to prioritize perturbations that are most likely to drive the observed phenotype in a given experimental context. While previous work has superficially explored whether machine learning models could predict phenotypic screens in silico~\citep{song2025virtual}, there is, to our knowledge, currently no widely adopted benchmark designed specifically for this task.

Despite its practical importance, progress toward in silico phenotypic screening remains difficult to measure systematically. Existing benchmarks for perturbation modeling focus on molecular readouts, primarily transcriptomic responses from single-cell assays~\citep{peidli2024scperturb,wu2024perturbench,wu2025contextualizing,youngblut2025scbasecount}, whereas phenotypic screens succeed or fail at the level of an assay endpoint, e.g., survival, proliferation, infection burden, reporter activity, or trafficking. The central prediction problem is therefore to \emph{prioritize the perturbations most likely to drive a functional phenotype in a specific experimental context}. While prior work has begun to explore this setting~\citep{song2025virtual}, no broadly adopted benchmark currently exists.

%On this work, we introduce \screensqa{}, a benchmark for phenotypic screen prediction derived from publicly available CRISPR screening data. \screensqa{} is built from BioGRID ORCS screens~\citep{oughtred2021biogrid} and organizes each screen as a ranking problem over genes, where the objective is to recover the perturbations most relevant to the measured phenotype. This framing is motivated by how screens are used in practice: researchers are often interested less in predicting a calibrated score for every gene than in obtaining a ranked list of candidates for follow-up. Because raw score semantics and hit definitions vary substantially across screens, we convert heterogeneous screen-specific significance criteria into a unified relevance representation and evaluate predictions with a ranking metric tailored to this setting. Specifically, we introduce the Adjusted normalized Discounted Cumulative Gain (AnDCG), a new metric based off nDCG~\citep{wang2013theoretical} that provides a continuous and comparable measure of performance across heterogeneous assays while accounting for screen-specific baselines. Our benchmarks comprises a total of 1920 screens from five general phenotype classes, with an average of 13,807 genes evaluated per screen.

In this work, we introduce \screensqa{}, a benchmark for phenotypic screen prediction built from publicly available CRISPR screens. \screensqa{} integrates screens from BioGRID ORCS~\citep{oughtred2021biogrid} and casts each screen as a gene-ranking problem conditioned on a free-text description of the screen.
The ranking formulation is more tractable than scoring each gene individually and mirrors how candidates are prioritized in real-world screening workflows. The free-text representation accommodates the diversity of readouts, experimental conditions, and ranking criteria, which would be difficult to capture in a fixed schema.
% This framing is more tractable than scoring each individual gene and more aligned with the drug discovery pipeline where researchers focus on prioritizing a list of candidates for downstream validation.
Because score semantics and hit definitions vary substantially across screens, we harmonize heterogeneous significance criteria into a unified notion of relevance. We leverage a temporal train/test splitting strategy to induce a realistic distribution shift. Furthermore, we evaluate predictions using a ranking metric tailored to this setting, introducing Adjusted normalized Discounted Cumulative Gain (\andcg), a variant of \ndcg~\citep{wang2013theoretical} that enables continuous evaluation across heterogeneous assays while correcting for screen-specific baselines. The resulting benchmark comprises  1{,}920 screens spanning five broad phenotype classes, with an average of 13{,}826 genes evaluated per screen.

While \screensqa{} is not intrinsically tied to any model class, in practice, its heterogeneous textual inputs and diverse phenotypic outputs make it especially well-suited to LLMs and agentic systems.
By contrast, most existing perturbation prediction models, such as GEARS~\citep{roohani2024predicting} or biolord~\citep{piran2024disentanglement}, including biological foundation models such as scGPT~\citep{cui2024scgpt}, are not readily applicable in this setting, as they are typically trained for specific modalities (most commonly gene expression data) and depend on structured, predefined input fields.
Therefore, \screensqa{} doubles as a testbed for evaluating LLMs as surrogates for virtual cells, providing fertile ground for future research in this area.

\screensqa{} complements existing benchmarks at the intersection of ML and biology.
Unlike molecular perturbation-response benchmarks~\citep{peidli2024scperturb, wu2024perturbench, wu2025contextualizing}, it targets screen-level phenotypic hit prediction across a heterogeneous range of assays.
Unlike image-based phenotypic profiling datasets~\citep{sypetkowski2023rxrx1}, which focus on morphological representations of physical assay data, it evaluates in silico perturbation prioritization.
Unlike LLM benchmarks for bioinformatics, biomedical knowledge, and question answering~\citep{mitchener2025bixbench, jiang2025benchmarking, rein2024gpqa, phan2025humanity}, \screensqa{} evaluates whether models can use biological context to predict experimentally measured phenotypic outcomes.

\textbf{Contributions.} (i) We introduce \screensqa{}, the first large-scale benchmark for phenotypic screen prediction, comprising 1{,}920 publicly available CRISPR screens that span five phenotype classes and use an assay-endpoint formulation to mimic real-world screening workflows. (ii) We formalize the task as screen-specific gene ranking under a temporal split and propose an evaluation protocol centered on adjusted \ndcg{} (\andcg{}), a continuous ranking metric corrected for screen-specific random baselines. (iii) We provide an extensive empirical study across frontier LLMs, biology-specific LLMs, trainable gene-relevance predictors, and retrieval/frequency baselines, and show that (a) no current method approaches the empirical ceiling, (b) off-the-shelf frontier  LLMs outperform biology-specific models and learned baselines, with evidence of scaling trends, and (c) optimization techniques such as supervised fine-tuning, prompt optimization, and learned ensembling can further improve LLM performance, supporting future research in this area.

%% file: sections/2data.tex
\section{Data preparation}

\begin{figure}
    \centering
    \includegraphics[width=\linewidth]{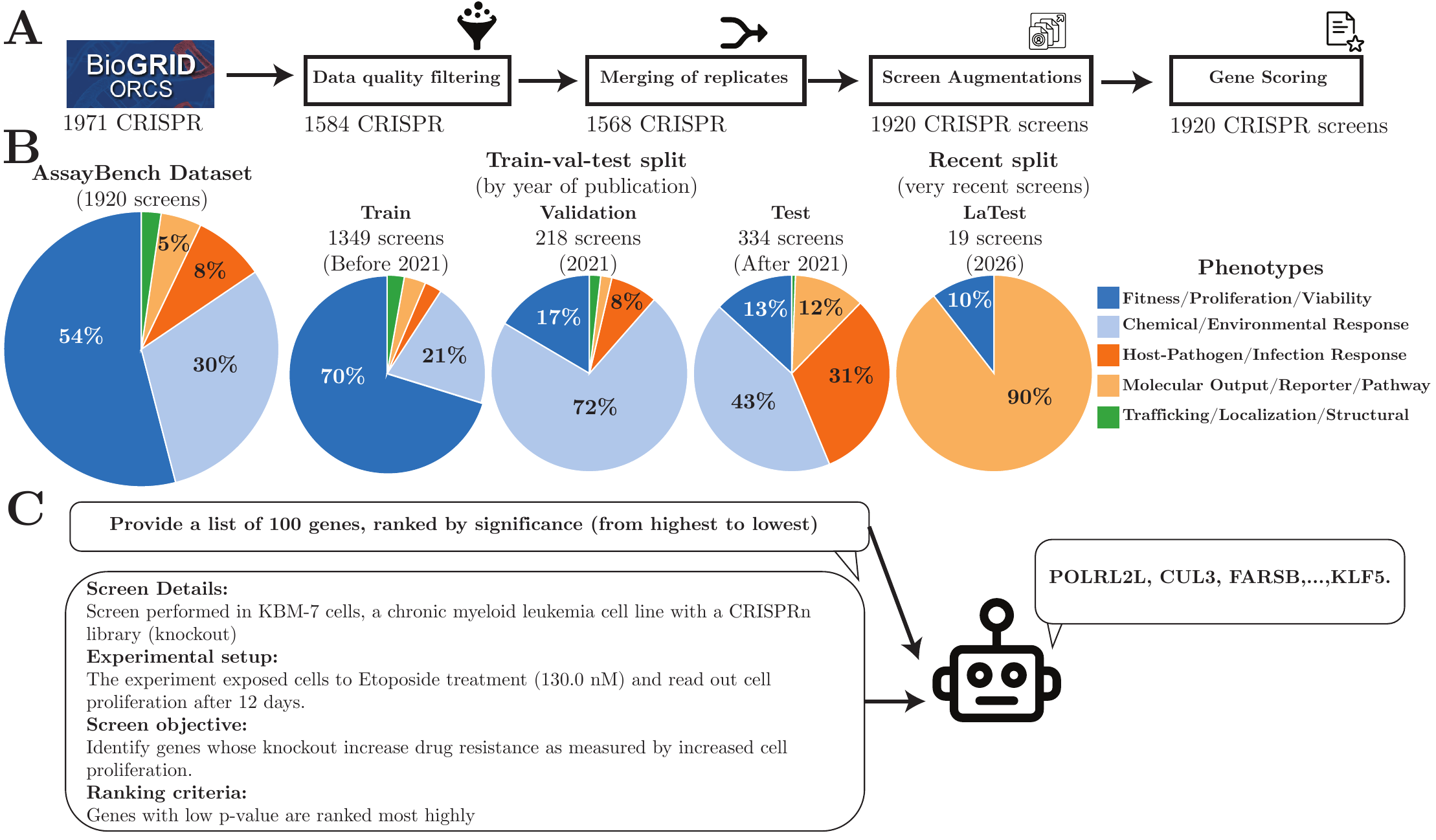}
    \caption{Overview of the \screensqa{} benchmark creation. (\textbf{A}) Starting from 1971 human CRISPR screens, we perform data quality filtering, replicate merging, and data augmentation to obtain 1920 high quality screens. (\textbf{B}) Phenotype  composition of the database and its four splits. A realistic but challenging temporal split was used. (\textbf{C}) Given a description of the screen and a gene ranking criteria, a model must provide a ranked list of 100 genes.}
    \label{fig:figure1}
\end{figure}

Figure~\ref{fig:figure1} summarizes the data curation pipeline. Starting from human CRISPR screens in BioGRID ORCS (1{,}952 screens) and recent publications (19 screens), our pipeline produces 1{,}920 phenotypic screening benchmark entries, each associated with a list of genes and their relevance score in the screen. We refer to the 19 screens sourced from recent literature and not available in BioGRID as the \novel{} split. Descriptive statistics are given in Table~\ref{tab:biogrid_dataset_stats}. The final dataset is available at \url{huggingface.co/datasets/Genentech/assaybench}.

\subsection{Screen curation and harmonization}
\label{sec:curation}

%Our main data sources are the BioGRID ORCS raw screen tables and screen-level index metadata~\citep{oughtred2021biogrid}. We used the 2025 version of the dataset that contains 1952 human CRISPR screens. Each BioGRID screens consists of (a) a screen description, (b) a significance criteria, and (c) a list of genes with the corresponding scores and whether each gene was a hit or not. We refer to a "hit" as a gene that satisfies the provided statistical significance criteria.

Our primary source is the 2025 BioGRID ORCS release~\citep{oughtred2021biogrid}, containing 1,952 human CRISPR screens. Each screen provides a textual experiment description and a significance criterion for calling hit genes. %, and a per-gene score table with hit calls.
Throughout the paper, we use the term \emph{hit} to denote a gene that satisfies the significance criterion of its screen.

We first remove screens that cannot support a meaningful ranking task, namely screens in which all tested genes are significant or with missing significance criteria. We then normalize gene symbols to HGNC nomenclature. % and collapse aliases mapping to the same approved HGNC symbol into a single gene entry.
Screens with identical metadata fields are treated as technical replicates and merged into a single benchmark entry. We identify 32 such replicate screens leading to 16 merged entries. %Details about gene replicates merging are given in Section~\ref{sec:scoring}.

Reported phenotype and the direction of the perturbation effect are often incomplete or ambiguous in BioGRID. We therefore use an LLM-assisted curation step that extracts, for each screen, a detailed phenotype description and an effect direction indicating whether perturbing a hit gene is expected to increase or decrease the measured phenotype. Phenotypes are further grouped into five broad categories (Figure~\ref{fig:figure1}B);
% : \textit{Fitness / Proliferation / Viability}, \textit{Drug / Chemical / Environmental Response}, \textit{Host-Pathogen / Infection Response}, \textit{Molecular Output / Reporter / Pathway Activity}, and \textit{Trafficking / Localization / Structural Phenotypes}. 
see Appendix~\ref{app:coarse_phenotype_procedure} for details.

% \subsection{Technical replicate merging}
% \label{sec:replicates}

% Screens with identical metadata fields are treated as technical replicates and merged into a single benchmark entry. We identify 32 replicate screens collapsing into 16 merged entries. Details about gene replicates merging are given in Section~\ref{sec:scoring}.

% \subsection{Screen augmentation}
% \label{sec:augmentations}

Some BioGRID screens use bidirectional significance criteria, in which genes are called hits if they either increase or decrease the measured phenotype. We decompose each such screen into up to three benchmark entries -- one per phenotype direction and an optional merged bidirectional entry -- expanding the 1{,}568 curated screens into 1{,}920 benchmark entries.
% When such a criterion can be cleanly decomposed into two opposite phenotype directions, we create one benchmark entry for each direction, leading up to three entries from a single screen (each direction and a merged bidirectional entry). 
%This procedure can yield up to three entries from a single BioGRID screen: one entry for each directional phenotype and, when appropriate, a merged bidirectional entry. 
%Augmented entries with no significant genes are discarded. 
% Overall, this step increases the dataset size from 1,549 curated screens to 1,901 benchmark entries.

\subsection{Gene relevance scoring}
\label{sec:scoring}

%A crucial part of our data curation process is assigning a relevance score to each gene in each screen. For a given screen, we start by identifying the list of scores present in the significance criteria $\{\mathbf{s}_k \in \mathbb{R}^G:k\in [K]\}$ where $K$ is the number of distinct scores in the significance criteria (\emph{e.g.} log fold change, or p-value) and $G$ is the number of genes tested in the screen. For each score, we infer its direction $\mathbf{d}_k\in \{0,1\}$ (whether higher or lower score is better for significance) from the significance criteria. We then compute the percentile of gene in each relevant score, according to the inferred direction, leading to a new vector of ranks $\mathbf{r}_k$ where higher is better. We then compute the relevance score $\tilde{\mathbf{y}}$ using a geometric mean of percentiles: $\tilde{\mathbf{y}} = \exp{\big(\frac{1}{K}\sum_{k=1}^K \frac{\mathbf{r}_k}{G}\big)}$. This construction ensures that the most significant genes have a relevance score close to 1, and least significant genes have a score close to 0. For technical replicates (Section~\ref{sec:replicates}), we take an additional geometric mean of the relevance scores from each individual screen.

A central step in the benchmark construction is assigning a continuous relevance score to each gene in a screen to serve as a proxy for its overall significance in the screen. Since a significance criterion may involve multiple metrics (e.g., both log-fold change and $p$-value), we let $\{\mathbf{s}_k \in \mathbb{R}^G : k \in [K]\}$ denote the $K$ vectors of per-gene metrics for a screen assessing $G$ genes. %and $G$ the number of assayed genes in the screen.
For each metric, we infer from the criterion whether larger or smaller values correspond to stronger significance. Each metric is then transformed into a percentile rank $\mathbf{p}_k \in [0,1]^G$ such that a larger value for a gene indicates stronger evidence of that gene's significance in the screen.

When the significance criterion involves multiple score columns ($K > 1$), we combine their percentile ranks using a geometric mean to obtain a preliminary relevance score $\tilde{\mathbf{y}}=\exp\!\left(
\frac{1}{K}
\sum_{k=1}^{K}
\log\!\big(\max(\mathbf{p}_k, 10^{-10})\big)
\right)$.
% This construction assigns values close to 1 to the most significant genes and values close to 0 to the least significant ones. %For merged technical replicates (Section~\ref{sec:replicates}), we apply an additional geometric mean across the replicate-specific relevance scores.
We assess the concordance of the ranking obtained from $\tilde{\mathbf{y}}$ and the original hit labels with ROC-AUC, excluding screens with ROC-AUC below 0.95. Finally, we set the relevance score of non-hit genes to zero, yielding a sparse final relevance vector $\mathbf{y} \in [0,1]^G$.

For screens with bidirectional significance criteria that we decomposed into multiple benchmark entries
 (Section~\ref{sec:curation}), we also leverage the distinction between genes associated with the target phenotype and genes associated with the opposite phenotype. Genes that are significant in the opposite direction are assigned negative relevance values, obtained by negating their directional score. As a result, these directional screens can contain positive, zero, and negative gene relevance scores.

%For bi-directional screens that were augmented into single-directional screen (Section~\ref{sec:augmentations}), we leverage the distinction between gene positively and negatively associated with the phenotype. Genes that significantly modulate the phenotype in the opposite direction are given a negative relevance score that is obtained by taking the opposite score of the gene in the screen with opposite direction.

\subsection{Temporal data splitting strategy}
\label{sec:splitting}
%Our main splitting strategy is a temporal split assigning screens published in or before 2020 to training, screens from 2021 to validation, and screens from 2022 onward to test. This split is intended to measure temporal generalization to newer screens and therefore induces a realistic distribution shift. As shown in Figure~\ref{fig:figure1}, this results in a vastly heterogenous phenotype composition across the three splits.

Our primary evaluation protocol uses a temporal split: screens published before 2021 are assigned to the training set, screens published in 2021 to the validation set, and screens published after 2021 to the test set. This split is intended to measure temporal generalization to novel screens and induces a realistic distribution shift. In the resulting split, the benchmark contains 1,349 training entries, 218 validation entries, and 334 test entries.

%Time since publication is also a crucial parameter when it comes to evaluate large language models whose training corpus encompass most of available public data. To further study generalization we created an additional collection of screens that were only very recently published (in the first quarter of 2026), which we refer to as \texttt{2026Q1}. We identified 19 screens from \edward{how many? + cite} such recent publications and processed them using the same procedure as for the BioGRID screens. Details for these screens are given in Appendix~\edward{TODO}.

Time since publication is also especially relevant when evaluating LLMs, whose pretraining corpora likely include much of the public literature used to construct the benchmark. To probe generalization, we assembled an additional test set of 19 screens from recent publications (after September 2025), which we refer to as \novel{}; see Appendix~\ref{app:2026Q1} for details.
% We identify 19 such screens from recent publications and process them using the same pipeline as the BioGRID-derived benchmark, leading to a total of 1{,}920 screens in our benchmark. Details are provided in Appendix~\ref{app:2026Q1}.

\subsection{Prompt generation}

For each screen, we generate a plain-text prompt that summarizes the experimental context, the significance criterion, and the ranking objective. The prompt includes the phenotype, cell line, cell type, CRISPR library and perturbation modality, experimental setup, treatment condition, and treatment duration (Figure~\ref{fig:figure1}(C), full prompt template in Appendix~\ref{app:prompt_template}).

 \input{tables/descriptive_statistics}

%% file: tables/descriptive_statistics.tex
\begin{table}
\caption{\screensqa{} dataset statistics by split.}
\label{tab:biogrid_dataset_stats}
\begin{adjustbox}{width=\linewidth}
\begin{tabular}{lccccc}
\toprule
 & Total & Train & Val & Test & \texttt{LaTest} \\
\midrule
Benchmark entries & 1,920 (100.0\%) & 1,349 (100.0\%) & 218 (100.0\%) & 334 (100.0\%) &  19 (100.0\%) \\
Number of unique screens (including merged replicates) & 1,584 & 1206 & 143 & 216 & 19 \\ 
Avg. number of tested genes per screen & 13,826 & 15,188 & 9,528 & 11,126 & 11,993 \\
Merged replicate entries & 16 (0.8\%) & 5 (0.4\%) & 9 (4.1\%) & 2 (0.6\%) & 0 \\
\hline
Phenotype: Drug / Chemical / Environmental Response & 590 (30.8\%) & 280 (20.7\%) & 166 (73.1\%) & 144 (42.9\%) & 0 \\
Phenotype: Fitness / Proliferation / Viability & 1,031 (53.7\%) & 949 (70.1\%) & 36 (15.9\%) & 44 (13.1\%) & 2 (10.5\%) \\
Phenotype: Host-Pathogen / Infection Response & 163 (8.5\%) & 39 (2.9\%) & 17 (7.5\%) & 107 (31.8\%) & 0 \\
Phenotype: Molecular Output / Reporter / Pathway Activity & 108 (4.7\%) & 48 (3.5\%) & 4 (1.8\%) & 39 (11.6\%) & 17 (89.5\%) \\
Phenotype: Trafficking / Localization / Structural Phenotypes & 44 (2.3\%) & 38 (2.8\%) & 4 (1.8\%) & 2 (0.6\%) & 0\\
\bottomrule
\end{tabular}
\end{adjustbox}
\end{table}

%% file: sections/3metrics.tex
\section{Task and evaluation metrics}

%Given a screen description and a gene ranking rationale, the objective of the model is to provide a ranked list of 100 genes that are most align with the provided rationale. To measure performance, we introduce a new metric, the adjusted normalized discounted cumulative gain (AnDCG\at{k}). We also report Precision\at{k} and the Opposite Phenotype Direction Rate (OPDR\at{k}), which we describe next.

\screensqa{} frames each screen as a ranking task: given the screen description and its ranking criterion, a model must produce a ranked list of 100 candidate genes. We evaluate predictions with three complementary metrics: Adjusted normalized Discounted Cumulative Gain (\andcg\at{k}), \Prec\at{k}, and directional False Discovery Rate (\fdr\at{k}), after canonicalizing each predicted list to HGNC symbols and removing duplicates.%; symbols not assayed in the target screen are mapped to a \texttt{MISSING} gene symbol.

% Given a screen description and its ranking criterion, the goal is to produce a ranked list of 100 genes that are most relevant to the phenotype of interest. We evaluate predictions using three complementary metrics: Adjusted normalized Discounted Cumulative Gain (AnDCG\at{k}), Precision\at{k}, and False Discovery Rate (FDR\at{k}).

% For all metrics, we first canonicalize the predicted gene list by mapping gene symbols to HGNC symbols and removing duplicates while preserving order. Predicted genes that were not assayed in the target screen are mapped to the \texttt{MISSING} gene symbol. %By default, invalid gene strings are treated the same way.

%For all metrics, we first map each gene to its HGCN symbol and remove duplicates. Predicted genes that do not map to a HGCN symbol or have not been measured in the screen are mapped to the \texttt{MISSING} gene symbol.

\subsection{\andcg\at{k}.}

Our primary metric, \andcg\at{k}, is a modified version of \ndcg\at{k}~\citep{wang2013theoretical} with two modifications: (i) it is \textbf{condensed}, ignoring unassayed genes rather than penalizing them as false positives, and (ii) it is \textbf{adjusted}, subtracting a screen-specific random baseline so that scores are comparable across heterogeneous assays.

% , which is more accurately described as an \emph{adjusted-condensed-normalized DCG\at{k}}. Compared with standard nDCG~\citep{wang2013theoretical}, this metric accounts for the fact that not all genes are measured in every screen (\emph{condensed}) and adjusts for a screen-specific random baseline (\emph{adjusted}), making scores more comparable across heterogeneous assays.

%Given a ranked list of genes $[g_0,...,g_N]$, we built the prediction score vector $\mathbf{x}$ such that $\mathbf{x}_i=\mathbf{y}[g_i]$. That is, at position $i$, the prediction score is the relevance score of the gene in the screen as given by the ground truth $\mathbf{y}$. We propagate the \texttt{MISSING} label to the predicted score: $\mathbf{y}[\texttt{MISSING}]=\texttt{MISSING}$. We then take the first $k$ elements of the list.

Let $[g_1,\dots,g_L]$ be a ranked prediction list, and let $\mathbf{y}$ denote the ground-truth relevance scores for the target screen. We construct the score sequence $\mathbf{x}$ by assigning
\[
x_i =
\begin{cases}
\mathbf{y}[g_i], & \text{if } g_i \text{ is assayed in the screen},\\
\texttt{MISSING}, & \text{otherwise}.
\end{cases}
\]
We then truncate this sequence to its first $k$ positions.

\textbf{Condensing step.} Because the set of assayed genes differs across screens, we do not treat unmeasured genes as ordinary false positives. Instead, after truncation to the top-$k$ positions, we remove all \texttt{MISSING} entries while preserving order. For example, $ [0.9,\,0.7,\,\texttt{MISSING},\,0.6]
\;\rightarrow\;
[0.9,\,0.7,\,0.6]$. We denote the condensed sequence by $\mathbf{x}'=(x'_1,\dots,x'_{k'})$, where $k' \le k$.

%Not all possible genes are tested in every single screen. To avoid penalizing predicted genes for which no ground truth is available, we first condense the predicted list by removing the \texttt{MISSING} tokens. \emph{E.g.,} $[0.9,0.7,\texttt{MISSING},0.6]\rightarrow[0.9,0.7,0.6]$. We write $\mathbf{x}'$ the result of the condensation of $\mathbf{x}$ and $k'$ the remaining number of elements in the list.

The \textbf{Discounted Cumulative Gain (\dcg\at{k})} is defined as $ \mathrm{DCG}@k = \sum_{i=1}^{k'} \frac{x'_i}{\log_2(i+1)}.$ Because relevance scores can be negative for genes associated with the opposite phenotype direction, ranking such genes near the top decreases the score.

\textbf{Ideal DCG and \ndcg.} The ideal ranking is obtained by sorting the ground-truth relevance scores in descending order. Let $\mathrm{IDCG}@k$ denote the corresponding discounted cumulative gain. We then define the \emph{normalized} DCG (\ndcg) as $ \mathrm{nDCG}@k =\frac{\mathrm{DCG}@k}{\mathrm{IDCG}@k}$ (and set it to zero when the denominator is zero).

\textbf{Adjusted \ndcg.} Raw \ndcg{} values are not directly comparable across screens because making predictions for some assays is intrinsically easier than for others. To correct for this, we compute a screen-specific random baseline $\mathrm{nDCG}_{\mathrm{rand}}@k$ and define
\begin{equation}
    \mathrm{AnDCG}@k = \frac{\mathrm{nDCG}@k - \mathrm{nDCG}_{\mathrm{rand}}@k}{1 - \mathrm{nDCG}_{\mathrm{rand}}@k}.
\end{equation}
In practice, this rescales performance so that 0 corresponds to a random ranking and 1 to the ideal ranking. Values below 0 are possible for predictors that perform worse than random. Notably, the expected value of the \ndcg\at{k} under a random predictor can be computed analytically, as shown in Appendix~\ref{app:adjusted_condensed_ndcg}.
%By design, $nDCG@k\leq1$. However, due to various levels of difficulty across screens, the dynamic range of $nDCG@k$ can vary widely over the dataset. This hinders performance comparison across screens. To address this issue, we first compute $nDCG_{rand}@k$, the expected value of $nDCG@k$ over a uniformly random distribution of predicted lists. We then compute our final metric as:

%\begin{equation}
%    AnDCG@k = \frac{nDCG@k - nDCG_{rand}@k}{1- nDCG_{rand}@k}.
%\end{equation}

%In practice, $AnDCG@k$ is bounded between 0 (random performance) and 1 (optimal performance). However, negative value could happen for a worse than random predictor.

\subsection{\Prec\at{k} and directional False Discovery Rate (\fdr\at{k})}

\Prec\at{k} is the fraction of top-k predictions that are hits: $\mathrm{Precision}@k = \frac{1}{\min(k,G^+)}\sum_{i=1}^k \mathbb{I}[\mathbf{x}_i>0]$, where $G^+$ is the number of positive-relevance genes in the screen. It captures the enrichment of true hit genes among the model's top in-screen predictions.
For directional screens which can assign negative relevance, we additionally report the directional False Discovery Rate: \fdr\at{k}$= \frac{1}{k} \sum_{i=1}^k \mathbb{I}[\mathbf{x}_i<0]$, the fraction of top-ranked scored predictions with negative relevance

% Some directional screens assign negative relevance to genes associated with the opposite phenotype direction (Section~\ref{sec:augmentations}). To measure this failure mode, we report the False Discovery Rate (FDR\atk), defined as the fraction of top-ranked scored predictions with negative relevance: $FDR@k = \frac{1}{k} \sum_{i=1}^k \mathbb{I}[\mathbf{x}_i<0]$.

\section{Benchmark models}

We compare a diverse set of modeling strategies, ranging from frontier large language models to trainable neural gene-level predictors and simple retrieval-based baselines.

\subsection{Large language models} 

% Large language models (LLMs) are a natural model class for \screensqa{} because the task requires integrating biological context, assay design, and phenotype semantics from text. 
We evaluate both closed-source frontier models---Gemini 3 Pro, Gemini 3 Flash~\citep{Gemini_2025}, and GPT-5.4~\citep{singh2025openai}---and open-weight models, including GPT-OSS-120B~\citep{agarwal2025gpt} and the Qwen3.5 family~\citep{qwen3.5}. Unless otherwise stated, these models are evaluated in a zero-shot setting.

%Being verstality and trained on large corpus of biology data, among others, large language models are a natural modeling class for \screensqa{}. We compared the performance of closed frontier language models - Gemini 3 Pro, Gemini 3 Flash~\citep{Gemini_2025}, and GPT 5.4~\citep{singh2025openai}; open-source models - GPT-OSS-120B~\citep{agarwal2025gpt}, and the Qwen3.5 suite (until 397B-A17B)~\citep{qwen3.5}. Unless otherwised mentioned, these models have been evaluated in a zero-shot fashion.

\textbf{Task-optimized LLMs.} Using the temporal training split (Section~\ref{sec:splitting}), we fine-tune GPT-OSS-120B with supervised fine-tuning (SFT), followed by reinforcement learning with GRPO~\citep{shao2024deepseekmath}. To adapt a larger proprietary model, we optimize prompts for Gemini 3 Flash using GEPA~\citep{agrawal2025gepa}. Finally, we evaluate an in-context learning variant of Gemini 3 Pro in which each test prompt is augmented with 10 training examples selected by nearest-neighbor retrieval over screen-description embeddings.

\textbf{Evolved ensemble approach.} We also explore automatically discovered ensembling strategies over LLM predictions, using AlphaEvolve~\citep{novikov2025alphaevolve} on the temporal training set to learn an algorithmic ensemble over a subset of model outputs (details in Appendix~\ref{app:ensemble}).

%On top of the above fine-tuning approaches, we explored code evolution strategies to automatically learn a ensembling strategy from different LLM predictions. In particular, we used AlphaEvolve~\citep{novikov2025alphaevolve} on the temporal training set to design an algorithm that optimally ensembles predictions from XXX, XXX \edward{@Carl can you confirm ? - TODO: decide name for the ensembling approach}.

\textbf{Biology-specific language models and agents.}
% Recognizing shortcomings of generalist language models in biological tasks, recent works have designed biology-specific language models and agents. 
We additionally evaluate the performance of C2S‑-Scale~\citep{rizvi2026scaling}, a language model fine-tuned on single-cell tasks, and Biomni~\citep{huang2025biomni}, a biomedical agent with access to biology-relevant tools.% (we use Claude Sonnet 4.5 as backbone model~\citep{anthropic2025claudesonnet45}).

\subsection{Trained neural gene-relevance predictors}
We train a neural gene-relevance predictor directly on the \screensqa{} training set. For each screen-gene pair, the model takes as input the concatenation of a text embedding of the screen description and a DeepSet aggregation~\citep{zaheer2017deep} over a collection of biologically informed gene embeddings, following prior work on gene-perturbation prediction \citep{littman2025gene}, and predicts a relevance score. At inference, we rank all candidate genes by predicted relevance. Architectural variants are reported in Appendix~\ref{app:relevance_predictor_classifiers}.

\subsection{Retrieval and frequency baselines}
\label{sec:TODO}
We evaluate three simple baselines. \texttt{Embedding kNN} predicts the top genes of the training screen most similar to the target screen, based on cosine similarity between screen-description embeddings. \texttt{Oracle kNN} selects the training screen maximizing \andcg\at{100} with respect to the target screen, serving as an upper bound on retrieval-based performance. \texttt{Gene-frequency} prioritizes genes by how often they are hits among training screens sharing the same phenotype category.

%% file: sections/4results.tex
\section{Results}

\input{tables/year_results_small}

\subsection{Frontier generalist LLMs lead benchmark performance}
\begin{figure}[h]
    \centering
    \includegraphics[width=\linewidth]{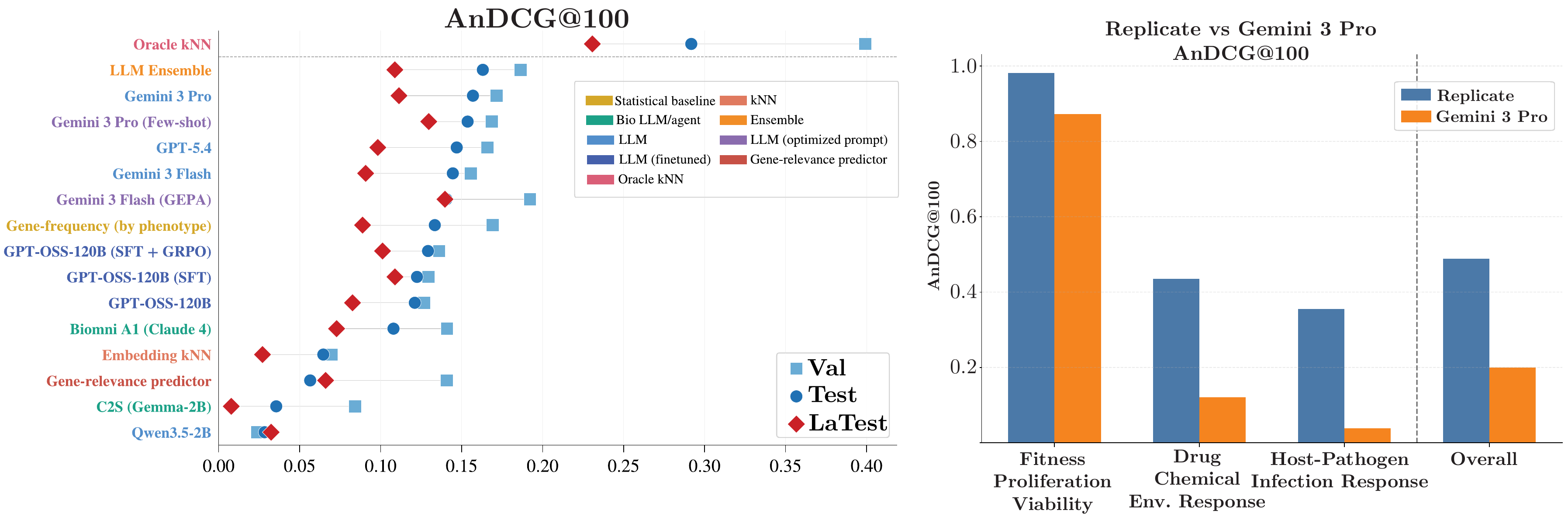}
    \caption{(Left) \andcg\at{k} on the main models, colored by model category. (Right) Comparison of \GeminiPro{} performance with a technical replicate baselines (\ref{sec:replicate_baseline}). ($N=32$ techincal replicate screens).}
    \label{fig:figure2}
\end{figure}

Figure~\ref{fig:figure2} and Table~\ref{tab:benchmark_results} summarize the main benchmark results. On the test set, Gemini 3 Pro and GPT-5.4 lead on \andcg\at{100}, outperforming smaller open-weight LLMs, biology-specific language models and agents (Biomni and C2S-Scale), the trainable neural gene-relevance predictor, and Embedding kNN.
Strong general-purpose LLMs are thus already competitive for assay-level phenotypic hit prediction, despite not being explicitly trained for this task.

%The comparison between \andcg\at{100} and \Prec\at{100} is also informative. The neural gene-relevance predictor achieves competitive precision on the test split, indicating that it often places true hits near the top, but its weaker performance on \andcg\at{100} suggests it is less effective at ordering those hits by graded relevance. In contrast, the strongest LLMs achieve better overall ranking quality, which is the primary objective of \screensqa{}. 
% This result also highlights the importance of the \andcg\at{100} metric in \screensqa{}.

% Another notable pattern is the degradation observed on the \texttt{2026Q1} screens. 
On the \novel{} screens, LLM-based methods suffer a larger drop than the neural gene-relevance predictor, consistent with the hypothesis that part of the advantage of frontier LLMs on older public screens may be partly driven by exposure to related literature during pretraining. This aspect is further investigated in Section~\ref{sec:memorization}.

The gene-frequency baseline is also surprisingly competitive overall. As we show in Section~\ref{sec:viability}, much of its signal is driven by the \textit{Fitness / Proliferation / Viability} phenotype subset, where frequently recurring essential or growth-associated genes create a strong phenotype-level prior. Full results on all splits are provided in Appendix~\ref{app:additional_results}. %The relative ranking of methods is broadly consistent across splits, though retrieval-based baselines benefit more from the random split, where train and test screens are drawn from overlapping time periods.

%Previous work had reported that model trained specifically on the phenotypic screen prediction could outperform frontier models~\citep{song2025virtual}. 

%The results of our benchmark are presented in Figure~\ref{fig:figure2}. On the test set, Gemini 3 Pro and GPT 5.4 clearly outperform smaller models like GPT-OSS-120B, biology-specific language models and agents like Biomni and C2S-Scale, as well as the trainable neural gene-relevance predictor and Embedding kNN in terms of \andcg\at{100}, our main metric. \Prec\at{100} preferred the neural gene-relevance predictor over GPT-5.4, suggesting a general ability at predicting screen hits, but lacking accurate ranking within them. Remarkably, while all LLM-based baselines incurred a significant loss in performance on the 2026Q1 dataset, the neural gene-relevance predictor performance remained comparable, suggesting some kind of memorization phenomenon in LLM that could hamper their real world performance. The phenotype-based hit frequency baseline was highly competitive, mostly explained by its performance on viability screens, as shown in Section~\ref{sec:viability}.

\subsection{Top-performing models remain far from the performance ceiling}
\label{sec:replicate_baseline}
Biology is inherently stochastic and experiments introduce further technical variability, raising the question of what performance ceiling can reasonably be attained by any model. We estimate this upper bound through two approaches.
First, the \texttt{Oracle kNN} shows that a model capable of perfectly retrieving the most relevant training screen would outperform the best model tested by 86\%.
Second, we use technical replicates (repetitions of the same experimental protocol, whose residual variability reflects irreducible biological and experimental stochasticity) identified in our data processing pipeline to design a replicate predictor that uses the list of top genes from one replicate to predict the other. As shown in Figure~\ref{fig:figure2} (right), this predictor nearly doubles the \andcg\at{100} of the best-performing LLM (Gemini 3 Pro), highlighting that accurately solving the \screensqa{} tasks remains well beyond the reach of existing state-of-the-art models. 

\subsection{LLM optimization is a promising direction to further improve performance}

We found that optimizing LLMs was generally helpful. SFT and GRPO improve GPT-OSS-120B's test \andcg\at{100} by 1\% and 7\%, respectively. GEPA improves the performance of Gemini 3 Flash on validation, but fails to generalize on the test set, suggesting overfitting. The ensembling strategy achieved the highest test performance, suggesting potential for further gains in this direction. These improvements are even more significant on the \novel{} screens, where SFT and GRPO improved the \andcg\at{100} by 32\% and 23\% respectively, and both few-shot and GEPA-optimized Gemini 3 Pro improved over the base model, with the latter achieving overall best performance.

\subsection{Performance varies across phenotypes and model sizes}

\label{sec:viability}

\begin{figure}[h]
    \centering
    \includegraphics[width=0.8\linewidth]{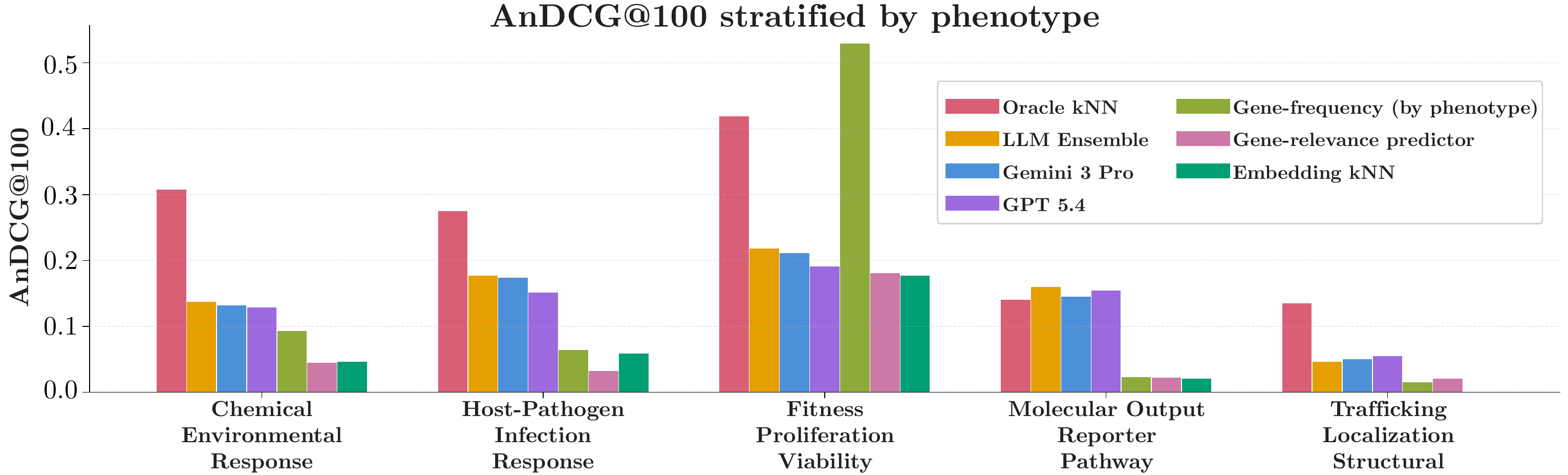}
    \caption{\andcg\at{k} of selected models stratified by phenotype on the test split.}
    \label{fig:figure3}
\end{figure}

Figure~\ref{fig:figure3} stratifies \andcg\at{100} of different models by phenotype. Predictive performance was highest for the viability screens, likely because their hit genes are enriched for conserved cellular dependencies that recur across screens. This recurrence also potentially explains why the phenotype-based frequency baseline is particularly strong for this class of screens.
% Viability screens are the most predictable class, likely because significant genes reflect conserved cellular dependencies that recur across screens, and the phenotype-based frequency baseline is strongest in this class.
% We note the high performance of the phenotype-based hit frequency baseline on the viability screens, explaining most of its overall performance. Other models likewise perform best on this phenotype. This suggests that viability screens are more predictable than other phenotype classes, likely because significant genes are more consistently conserved across screens.
% This result is biologically plausible. Viability-related screens often reflect shared cellular dependencies, so the same genes are repeatedly recovered across assays and contexts. 
% As a consequence, models can perform well by exploiting broad priors over gene essentiality or growth control, even without capturing the full details of the screen.
Other phenotypes, such as host-pathogen response or molecular reporter activity, appear more context-specific and therefore harder to predict from generic biological knowledge alone.
%Additionally, we observe how for molecular output screens, the kNN-based approach leads to a significantly lower upper bound, with multiple models outperforming it.

% \subsection{Scaling trends}

We also examine scaling within the Qwen 3.5 family (Figure~\ref{fig:figure4}, left). Larger models generally achieve higher \andcg\at{100}, consistent with a scaling trend, with gains plateauing at the top end, possibly because the 35B, 122B, and 397B variants are mixture-of-experts (MoE) models, whose active parameter counts are smaller than their total parameter counts.
% indicating a positive scaling trend on \screensqa{}. This trend suggests that assay-level phenotypic prediction benefits from the broader knowledge and reasoning capacity of larger language models.

% At the same time, the gains begin to plateau at the upper end of the size range. One possible explanation is architectural: the 35B, 122B, and 397B variants are mixture-of-experts models, whose active parameter counts are smaller than their total parameter counts. %More broadly, scaling alone is unlikely to solve the task. The gap to oracle retrieval and replicate agreement suggests that better use of biological context and better handling of screen-specific structure will be necessary in addition to larger models.

\subsection{Evidence for memorization}
\label{sec:memorization}

\begin{figure}[h]
    \centering
    \includegraphics[width=\linewidth]{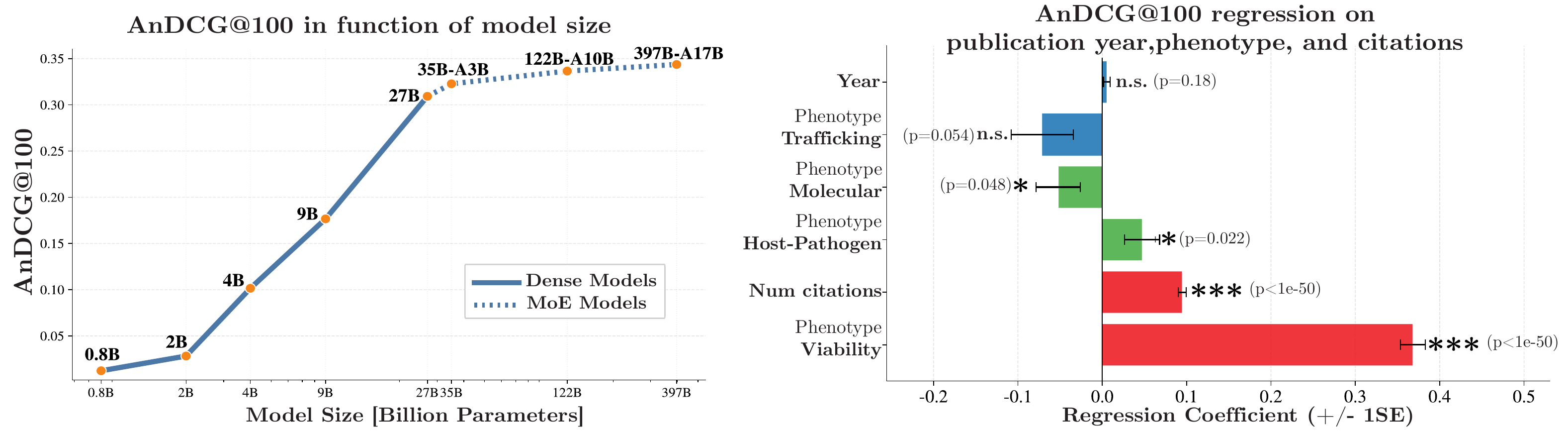}
    \caption{(Left) Scaling trend analysis on the Qwen3.5 family. Larger models lead to higher \andcg\at{k} performance. Larger models are mixture of experts (MoE), with a limited number of active parameters. (Right) Impact of different screen covariates on the performance of \GeminiPro{}. Number of citations is highly significant.\vspace{-5mm}}
    \label{fig:figure4}
\end{figure}

The performance drop on \novel{} motivates a closer look at possible memorization. A regression analysis of Gemini 3 Pro performance across \screensqa{} as a function of screen publication year, citation count, and phenotype shows the apparent temporal effect is largely explained by citations (Figure~\ref{fig:figure4}, Right).
% Notably, the apparent temporal effect is largely explained by citation count: once citation count is included in the model, publication year no longer appears to drive performance independently, whereas citation count remains strongly associated with \andcg\at{100}. 
This is consistent with memorization, as highly cited screens are more likely to have been discussed in the literature, increasing the chance that their biological findings were present in the pretraining data of frontier models.
Nonetheless, the fact that frontier LLMs retain the strongest relative performance on \novel{} suggests that their advantage is not solely driven by memorization. %, and further underscores the importance of the recent-screens subset of \screensqa{}.
This is further supported by performance gains across multiple generations of similarly-sized Qwen models (see  \ref{sec:gains_over_time}). 

%The degraded performance of LLMs on the 2026Q1 dataset observed in Figure~\ref{fig:figure2} hints at some memorization phenomenon. We further investigate this hypothesis by testing the impact of screen publication year and number citations on the performance of Gemini 3 Pro over the whole \screensqa{} dataset. We fit an ordinary least square regression with \andcg\at{100} as the response variable and the publication year, number of citations, and phenotype as predictors. We report the estimates for the contribution of each variable to the performance in Figure~\ref{fig:figure4}. Notably, all the temporal variation of the performance was explained by the number of citations (p-value$<1e-70$). Large number of citations entail a broader dissemination of a screen's results in the scientific literature, and a higher likelihood of being part of the LLM training corpus. 

\subsection{Evaluating biological biases in language models}

\begin{wrapfigure}{r}{0.35\textwidth}
    \vspace{-1.5\baselineskip}
    \centering
    \includegraphics[width=\linewidth]{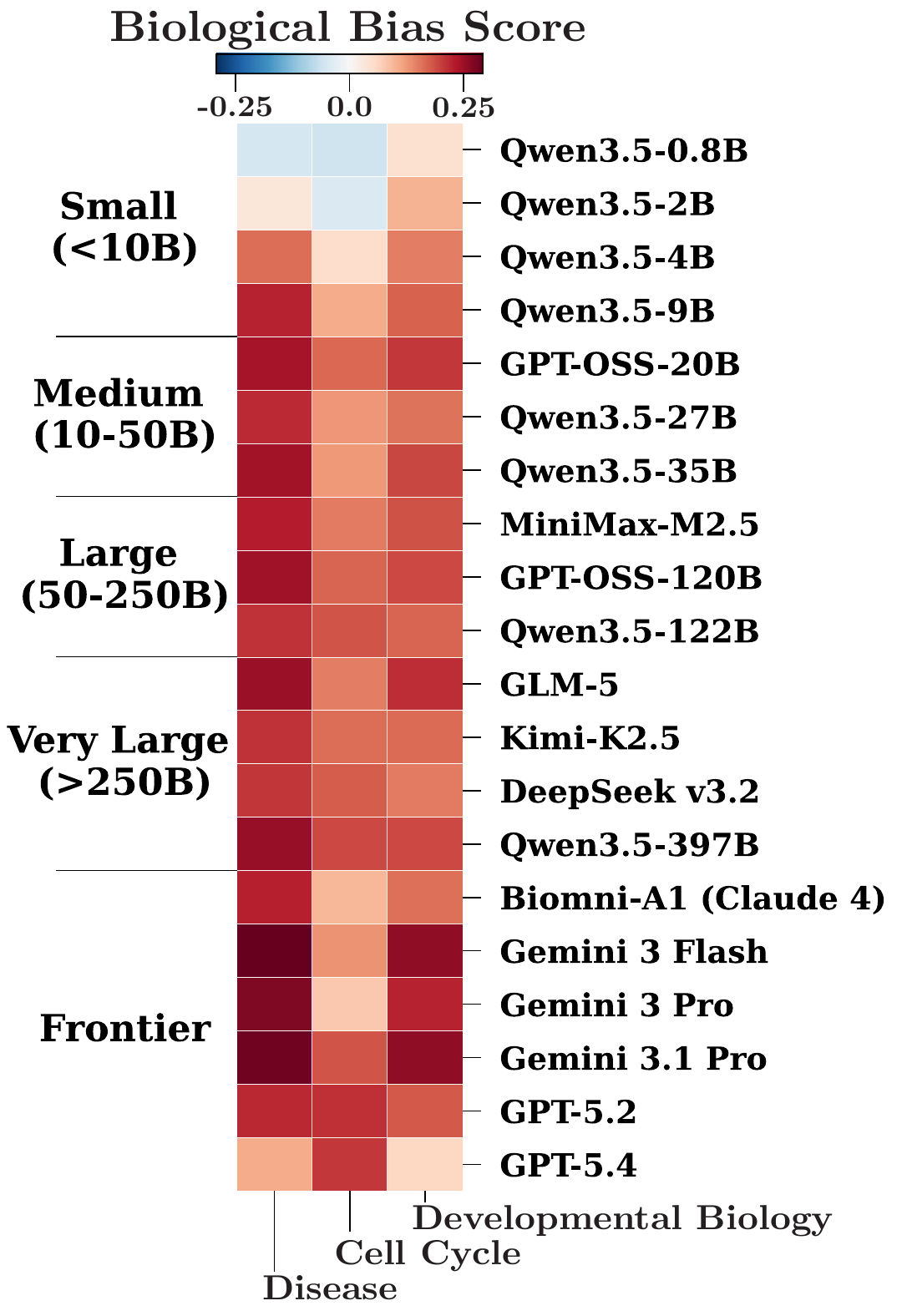}
    \caption{Biological bias of different models across gene sets.}
    \label{fig:biobias}
    \vspace{-2\baselineskip}
\end{wrapfigure}

To characterize model-specific biases, we compute for each (model, screen) pair the fraction of top-100 predicted genes belonging to several curated gene sets, minus the same fraction among ground-truth hits (Figure~\ref{fig:biobias}). A positive value indicates that the model over-represents genes from that set relative to the true screen outcome, while a negative value indicates under-representation. Systematic differences emerge between model families: GPT models over-represent cell-cycle genes, Gemini models over-represent developmental-biology genes, and all models over-represent disease-associated genes, likely reflecting the prevalence of those genes in the training corpus.

% To characterize the biological biases of different models, we analyze the composition of their top-100 predicted genes across screens with respect to curated gene sets. For each model and each screen, we compute the fraction of predicted genes belonging to a given gene set and subtract the corresponding fraction among the ground-truth hit genes. 
% A positive value indicates that the model over-represents genes from that set relative to the true screen outcome, while a negative value indicates under-representation. 
% This difference, averaged across screens, serves as a measure of biological bias and is reported in Figure~\ref{fig:biobias}. 
% The analysis reveals systematic differences between model families: for instance, GPT models tend to focus more on genes involved in cell cycle, while Gemini models appear to favor developmental biology-related genes. Overall, all models seem to be heavily biased by genes that are strongly associated with diseases, which could be explained by the larger prevalence of these genes in the literature and in the training corpus.

%% file: tables/year_results_small.tex
% Requires \usepackage{booktabs}
\begin{table}[t]
\centering
\small
\begin{tabular}{llrrr}
\toprule
\textbf{Method} & \textbf{\andcg\at{100}}$(\uparrow)$ & \textbf{\Prec\at{100}}$(\uparrow)$ & \textbf{\fdr\at{100}}$(\downarrow)$ \\
\midrule
Oracle kNN &  0.2918 & 0.3798 & 0.0037 \\
\midrule
Gemini 3 Pro   & 0.1570 & 0.2226 & 0.0164 \\
Gemini 3 Flash   & 0.1446 & 0.2009 & 0.0180 \\
GPT-5.4   & 0.1470 & 0.1980 & 0.0206 \\
Qwen3.5-2B   & 0.0284 & 0.0755 & 0.0216 \\
\midrule
LLM Ensemble   & 0.1631 & 0.2196 & 0.0178 \\
\midrule
Gemini 3 Pro (Few-shot)   & 0.1537 & 0.2134 & 0.0213 \\
Gemini 3 Flash (GEPA)   & 0.1406 & 0.1987 & 0.0214 \\
\midrule
GPT-OSS-120B   & 0.1211 & 0.1757 & 0.0223 \\
    +SFT   & 0.1224 & 0.1738 & 0.0255 \\
    +SFT+GRPO   & 0.1293 & 0.1699 & 0.0252 \\
\midrule
Biomni A1 (Claude 4)   & 0.1079 & 0.1701 & 0.0176 \\
C2S (Gemma-2B)   & 0.0355 & 0.0863 & 0.0218 \\
\midrule
Gene-relevance predictor   & 0.0565 & 0.1577 & 0.0306 \\
Embedding kNN   & 0.0646 & 0.1062 & 0.0226 \\
\midrule
Gene-frequency (by phenotype)   & 0.1334 & 0.2204 & 0.0292 \\
\bottomrule
\end{tabular}
\caption{Test split results for \andcg\at{100}, \Prec\at{100}, and \fdr\at{100}.}
\label{tab:benchmark_results}
\end{table}

%% file: sections/5discussion.tex
\section{Discussion}

\screensqa is, to our knowledge, the first large-scale benchmark for phenotypic screen prediction. It also provides a testbed for evaluating LLMs and agents as surrogates for virtual cells, supporting progress in this area. A key design choice is to cast each assay as a single ranking problem rather than issuing one query per gene: with an average of 13,826 genes per screen, gene-level evaluation would require over 20 million queries, making benchmarking impractical. The ranking formulation also better reflects real screening workflows, in which scientists prioritize a short list of candidates for follow-up experimental validation. More broadly, high performance on \screensqa requires integrating knowledge across gene function, pathway structure, cellular context, and assay design, making it not only a benchmark for phenotypic screening but also a test of whether a system can link mechanistic biology to experimentally measured phenotypes.

Our results show that current models remain far below the empirical ceiling. At the same time, both scaling trends within the Qwen family and the gains from fine-tuning GPT-OSS suggest that larger and better-adapted models can continue to improve. However, fine-tuning on \screensqa alone is limited by its modest size. A natural next step is identifying auxiliary tasks such as perturbation reasoning, pathway-level inference, and context-specific response prediction, and transferring that knowledge to \screensqa. Further, studying data scaling laws could help estimate the extra quantity of screen data needed to reach desired performance on arbitrary screens. Notably, our empirical ceiling estimate is itself conservative. Because BioGRID does not explicitly annotate replicates, we infer them by matching metadata fields; if some matched pairs are not true replicates, the gap to current models is even larger than reported. 

An important caveat is data memorization. Because the screens and associated publications are public, frontier LLM pretraining corpora likely include related findings, consistent with the strong association between citation count and model performance (Section~\ref{sec:memorization}). % and the drop on \novel screens.
However, the strongest relative performance on the \novel subset suggests that the performance of frontier LLMs is not solely driven by memorization. To preserve benchmark utility as models improve, we plan to update \screensqa quarterly with newly published screens, ensuring future versions continue to measure out-of-distribution generalization. % Relatedly, the public availability of the data complicates evaluating agentic systems: agents with web or database access could retrieve BioGRID records and bypass the intended reasoning task. We therefore exclude such systems when they cannot be reliably constrained from accessing ground-truth sources. Developing leakage-resistant evaluation protocols for tool-using biological agents remains an open problem.

%% file: sections/appendix/A_dataprep.tex
\section{Data processing}

\subsection{Benchmark creation}

We start from the 2025 BioGRID ORCS release~\citep{oughtred2021biogrid}, which provides a screen-level metadata index and per-screen gene tables containing gene-level hit labels and one or more numeric score columns for all assayed genes.

\paragraph{Initial cleaning and exclusion.}
Metadata fields with missing values are standardized to ``Not specified.'' Screens are excluded if all tested genes are marked significant (no meaningful foreground--background separation) or if the significance criterion is unparseable. Before ranking, a small number of systematically misannotated score types are corrected by inspecting the threshold and inferred significance direction (e.g., a field labeled ``p-Value'' with threshold $>0.5$ and positive direction is reinterpreted as ``$-\log$(p-Value)'').

\paragraph{Parsing significance rules.}
Each screen's significance criterion is parsed into elementary conditions of the form \emph{score $\{<,\leq,>,\geq\}$ threshold}. For each referenced score, the parser infers a direction: ``positive'' (higher values more significant), ``negative'' (lower values more significant), or ``bidirectional'' (the same score appears with both lower-tail and upper-tail thresholds). This parsed representation is used both to rank genes and to generate a human-readable ranking rationale.

\paragraph{LLM-assisted interpretation of phenotype direction.}
A separate LLM-assisted step determines how gene perturbation affects the screen phenotype. Three cases are distinguished.
For \emph{unidirectional} screens, the annotation specifies whether perturbing hit genes increases or decreases the phenotype; these remain single benchmark entries.
For explicitly \emph{bidirectional} screens, two phenotype/rule pairs (one per direction) are provided, and the screen is split into two directional benchmark entries.
If a screen cannot be classified as unidirectional, a fallback step attempts to identify a score whose sign distinguishes opposite phenotypic effects (``recovered bidirectional''). These screens are kept as a single entry with a generic phenotype description.

\paragraph{Within-screen gene ranking.}
Hit labels are taken directly from BioGRID. Ranking is computed over all genes with valid values in every referenced score column. Each score is transformed into a within-screen percentile (as described in Section~\ref{sec:scoring}), and when multiple scores are referenced they are combined via geometric mean. Genes are ranked by the resulting combined score in descending order.

\paragraph{Quality control.}
The combined score is evaluated against the BioGRID hit labels using AUROC as a consistency check. Screens with $\mathrm{AUROC} < 0.95$ are excluded. Screens with no hit/non-hit variation are discarded. For bidirectional screens, the threshold is applied independently to each directional branch.

\paragraph{Gene-symbol normalization.}
Gene symbols are mapped to official HGNC nomenclature using a mapper that resolves approved symbols, previous symbols, aliases, and UniProt-based protein-to-gene mappings. Unmappable symbols are dropped. When multiple original symbols map to the same HGNC symbol, they are collapsed: scores are averaged and the hit label is set to the logical OR across duplicates.

\paragraph{Relevance label construction.}
For standard screens, each gene receives a relevance score equal to its combined score if it is a hit, and zero otherwise. For directional entries derived from bidirectional screens, genes that are hits in the opposite direction receive \emph{negative} relevance scores. Genes called as hits in both directions are removed as contradictory. When both directional branches pass quality control, a third ``merged'' entry is also created, whose relevance label is the positive union of both branches.

\paragraph{Technical replicate merging.}
Technical replicates are identified from a precomputed duplicate annotation and merged on the intersection of their gene sets. For each common gene, the merged combined score is the geometric mean across replicates, raw scores are arithmetically averaged, and the hit label is determined by majority vote. Merged entries are retained only if they pass the same AUROC threshold.

\paragraph{Split assignment.}
Each benchmark entry is assigned to train, validation, or test under multiple splitting schemes (by publication year, random, author, cell line, and phenotype), each with three folds. For entries spanning multiple screen IDs (e.g., merged replicates), split assignment is conservative: if any underlying screen belongs to the test set, the merged entry is assigned to test.

\subsection{Derivation of coarse phenotype labels}
\label{app:coarse_phenotype_procedure}

Coarse phenotype labels were assigned to the 1{,}584 unique screens in the final benchmark using an LLM-assisted workflow followed by manual review.

\paragraph{LLM-based classification.}
For each screen, the full BioGRID metadata record (18 fields including analysis type, cell line, phenotype, library, and experimental conditions) was serialized to JSON and passed to GPT-5. The prompt asked the model to assign (i) exactly one primary phenotype category from a predefined list of nine candidates, (ii) a primary readout category, (iii) a normalized endpoint description, and (iv) a readout entity. The prompt instructed the model to prioritize assay design over topic words and to return structured JSON. The initial set of nine phenotype categories included fine-grained classes such as \texttt{Cell Cycle / DNA Damage / Genome Maintenance} and \texttt{Cell Death / Stress / Senescence}, which were later consolidated (see below).

\paragraph{Manual consolidation.}
After the LLM pass, the authors inspected screens assigned to rare or ambiguous categories. Fifteen screens were manually reassigned (e.g., eight phagocytosis-related screens were moved to \texttt{Morphology / Organelle / Trafficking / Localization}). The nine initial categories were then consolidated into the five coarse phenotypes used in the paper by merging related classes and renaming two categories:
\begin{align*}
&\texttt{Morphology / Organelle / Trafficking / Localization} \\
 &\quad \rightarrow \texttt{Trafficking / Localization / Structural Phenotypes}
\end{align*}
and
\begin{align*}    
&\texttt{Signaling / Reporter / Pathway Activity}\\
&\quad \;\rightarrow\;
\texttt{Molecular Output / Reporter / Pathway Activity}.
\end{align*}

The final phenotype distribution across the 1{,}565 benchmark screens is shown in Figure~\ref{fig:figure1}B.

\subsection{\novel{} dataset creation}
\label{app:2026Q1}

We used five recent publications (published between September 2025 and April 2026) where new CRISPR screens were publicly released.  \citet{bradu2026genome} use genome-wide single-cell CRISPRi Perturb-seq, with transcriptomic profiles and optional phenotypic enrichment as the main readout. \citet{zhu2025genome}  perform genome-scale Perturb-seq in primary human CD4+ T cells, measuring single-cell gene-expression changes in resting versus stimulated states to map regulators of T cell programs. \citet{jung2025virus} report pooled CRISPR knockout screens in primary human myeloid cells, with inflammatory macrophage phenotypes such as TNF production and CD80 expression as the readouts. \citet{burrell2026rational} use a genome-scale CRISPRa screen to identify factors that improve precise genome editing, with homology-directed repair efficiency as the phenotype. \citet{datlinger2025systematic} perform genome-wide CRISPR knockout screens in primary CAR T cells, measuring multiple therapeutic phenotypes including proliferation, target-cell killing/recognition, activation, apoptosis, fratricide, exhaustion, and in vivo antitumor efficacy.

\subsection{Prompt Template}
\label{app:prompt_template}

We provide the prompt template used to evaluate the LLMs below.

\begin{lstlisting}[style=promptstyle]
## Goal

You are tasked with ranking genes from a genetic perturbation screen. Based on the experimental context and hit criteria provided below, provide a list of 100 genes that are hits in this screen, ranked from strongest to weakest according to the criteria defined below.

## Experimental Context

This screen was performed in {{cell_line}} cells, a {{cell_type}}. Researchers used a {{library_type}} library ({{library_methodology}}) to systematically perturb gene function. The experiment followed a {{experimental_setup}} design and was conducted over {{duration}}{{condition_clause}}.

## Screen Objective

The primary objective of this screen was to identify a set of hit genes, each of which {{phenotype}}.

## Hit Definition

A gene is classified as a "hit" if its {{library_methodology}} significantly {{phenotype}}. The statistical criterion for significance is: {{significance_criteria}}.

## Ranking Criteria

Genes with {{ranking_rationale}} are ranked most highly.

## Additional Context

Screen notes: {{notes}}

## Required Output Format

Provide your response as an ordered list of exactly 100 HGNC gene symbols, using the ranking criteria above.
That is, top genes should have {{ranking_rationale}}.

Format:
GENE1, GENE2, GENE3, ..., GENE100
\end{lstlisting}

%% file: sections/appendix/B_metrics.tex
\section{Metrics}

This appendix provides the full specification of the evaluation metrics used in \screensqa{}. All metrics share a common preprocessing pipeline described below, followed by metric-specific computation.

\subsection{Common preprocessing}
\label{app:metric_preprocessing}

For one benchmark instance (one screen), let $\mathcal{G} = (g_1,\dots,g_N)$ be the dataset gene list and $\mathcal{R} = (r_1,\dots,r_N)$ the associated ground-truth relevance scores. Relevance scores may be positive (hit genes), zero (non-hits), or negative (genes associated with the opposite phenotype direction). A model produces an ordered prediction list $\mathcal{P} = (p_1,\dots,p_L)$.

\paragraph{Canonicalization.}
The raw model output is first parsed into a gene list, de-duplicated (preserving first occurrence), normalized to HGNC symbols, and de-duplicated again (since multiple raw strings may map to the same HGNC symbol).

\paragraph{Relevance assignment.}
Each predicted gene is mapped to a value:
\[
v_i =
\begin{cases}
r(g) & \text{if } g \text{ is present in the screen gene list},\\[4pt]
\texttt{None} & \text{otherwise (valid HGNC gene not assayed, or invalid symbol)}.
\end{cases}
\]
Genes assigned \texttt{None} are \emph{unscored} and will be removed by a condensation step. They do not contribute positive, zero, or negative gain to any metric. This design reflects the fact that a model typically cannot know which genes were assayed in a particular screen, so out-of-screen predictions should not be penalized as false positives.

\paragraph{Condensation.}
Condensation removes all \texttt{None} entries from a sequence while preserving the relative order of scored entries. The two metric families differ in \emph{when} condensation is applied relative to the top-$k$ cutoff (see below).

\subsection{Adjusted-Condensed-Normalized \dcg\at{k} (\andcg\at{k})}
\label{app:adjusted_condensed_ndcg}

\paragraph{Step 1: truncation, padding, and condensing.}
Let $\mathbf{v} = (v_1,\dots,v_L)$ be the relevance sequence. To compute the metric at cutoff $k$:
\begin{enumerate}
\item If $L < k$, append zeros until the list has length $k$.
\item Truncate to the first $k$ positions.
\item Remove all \texttt{None} entries (condensation).
\end{enumerate}
Crucially, condensation happens \emph{after} truncation: out-of-screen genes in positions $1,\dots,k$ are dropped, but genes ranked below $k$ do \emph{not} move up to replace them. For example, if
\[
\mathbf{v} = [1.0,\ \texttt{None},\ 0.3,\ \texttt{None},\ -0.2,\ 0.8]
\]
and $k=5$, the condensed sequence entering DCG is $[1.0,\ 0.3,\ -0.2]$, not $[1.0,\ 0.3,\ -0.2,\ 0.8]$.

Let $(c_1,\dots,c_m)$ denote the condensed sequence, where $m \le k$.

\paragraph{Step 2: DCG and normalization.}
The discounted cumulative gain is
\[
\mathrm{DCG}@k = \sum_{j=1}^{m} \frac{c_j}{\log_2(j+1)}.
\]
Because negative relevance values are retained, ranking oppositely relevant genes near the top decreases the score.

The ideal ranking is constructed by clipping negative relevance values to zero and sorting in descending order:
\[
\mathrm{IDCG}@k = \mathrm{DCG}@k\!\big(\mathrm{sort}(\max(\mathcal{R},0), \text{desc.})\big).
\]
Clipping ensures that the ideal predictor is not penalized for the existence of negatively relevant genes, preventing $\mathrm{nDCG} > 1$. The normalized score is $\mathrm{nDCG}@k = \mathrm{DCG}@k \,/\, \mathrm{IDCG}@k$ (set to zero when $\mathrm{IDCG}@k = 0$).

\paragraph{Step 3: adjustment.}
Raw $\mathrm{nDCG}$ values are not comparable across screens because some assays are intrinsically easier. We adjust relative to a screen-specific random baseline, derived as follows.

Consider a predictor that outputs a uniformly random permutation of the $N$ screen genes. Since every predicted gene belongs to the screen, no entries are unscored, and the condensed list after truncation has length $m = \min(k, N)$. At any position $j$, each gene is equally likely to appear, so the expected relevance is $\mathbb{E}[c_j] = \bar{r} = \frac{1}{N}\sum_{i=1}^{N} r_i$. By linearity of expectation,
\[
\mathbb{E}[\mathrm{DCG}@k] = \sum_{j=1}^{m} \frac{\mathbb{E}[c_j]}{\log_2(j+1)} = \sum_{j=1}^{m} \frac{\bar{r}}{\log_2(j+1)} = \mathrm{DCG}@k\!\big((\bar{r},\dots,\bar{r})\big).
\]
Since $\mathrm{IDCG}@k$ depends only on the ground truth and is independent of the predicted ranking, dividing both sides by $\mathrm{IDCG}@k$ gives
\[
\mathrm{nDCG}_{\mathrm{rand}}@k = \mathbb{E}[\mathrm{nDCG}@k] = \mathrm{nDCG}@k\!\big((\bar{r},\dots,\bar{r}),\, \mathcal{R}\big).
\]
The final metric is
\begin{equation}
\mathrm{AnDCG}@k = \max\!\left(\frac{\mathrm{nDCG}@k - \mathrm{nDCG}_{\mathrm{rand}}@k}{1 - \mathrm{nDCG}_{\mathrm{rand}}@k},\; 0\right).
\end{equation}
A value of $1$ corresponds to ideal ranking, $0$ to random-level or worse performance.

\subsection{Normalized \Prec\at{k} and \fdr\at{k}}
\label{app:precision_metrics}

Unlike the DCG-based metric, the precision-style metrics apply condensation \emph{before} the top-$k$ cutoff. This means that unscored predictions are first discarded, and the top-$k$ list is then taken from the remaining scored subset. Consequently, genes ranked below position $k$ in the original list can enter the evaluated set if higher-ranked predictions were unscored.

Concretely, if $\mathbf{v} = [1.0,\,\texttt{None},\,0.0,\,-0.7,\,\texttt{None},\,0.8]$, the condensed sequence is $[1.0,\,0.0,\,-0.7,\,0.8]$, and with $k=3$ the evaluated set is $[1.0,\,0.0,\,-0.7]$.

Let $(c_1,\dots,c_M)$ be the condensed sequence and $k' = \min(k,M)$.

\paragraph{\Prec\at{k}.}
\[
\mathrm{Precision}@k
=
\begin{cases}
\dfrac{1}{k'} \sum_{j=1}^{k'} \mathbf{1}[c_j > 0], & k' > 0,\\[8pt]
0, & k' = 0.
\end{cases}
\]
This measures the fraction of top-$k$ scored predictions that correspond to positively relevant genes.

\paragraph{\fdr\at{k}.}
Analogously,
\[
\mathrm{FDR}@k
=
\begin{cases}
\dfrac{1}{k'} \sum_{j=1}^{k'} \mathbf{1}[c_j < 0], & k' > 0,\\[8pt]
0, & k' = 0.
\end{cases}
\]
This measures the fraction of top-$k$ scored predictions with negative relevance, i.e., genes associated with the opposite phenotype direction. A high \fdr\at{k} indicates that the model is placing oppositely relevant genes near the top of its list.

\paragraph{Normalization.}
Some screens contain fewer than $k$ hits, yielding a maximum attainable precision below $1$. To enable cross-screen comparison, we normalize both metrics by their screen-specific maximum:
\[
\mathrm{NormPrecision}@k
=
\frac{\mathrm{Precision}@k}{\max_{R \subseteq \mathcal{G},\, |R|=k} \mathrm{Precision}@k(R)},
\qquad
\mathrm{NormFDR}@k
=
\frac{\mathrm{FDR}@k}{\max_{R \subseteq \mathcal{G},\, |R|=k} \mathrm{FDR}@k(R)}.
\]

%% file: sections/appendix/C_benchmarks.tex
\section{Evaluated models}

\subsection{Neural gene-relevance predictors}
\label{app:relevance_predictor_classifiers}

The gene-relevance predictor operates at the \emph{gene-within-screen} level: each screen is expanded into one training example per assayed gene, so that a single data point maps a (screen context, gene) pair to a continuous relevance score. At inference, the model scores all candidate genes for a given screen and ranks them in descending order.

\paragraph{Screen representation.}
The screen context is encoded as the concatenation of a text embedding and a direction embedding. The screen description is embedded with \texttt{text-embedding-3-small} (1536 dimensions). A four-dimensional one-hot vector encodes the phenotype direction (\emph{increases}, \emph{decreases}, \emph{increases or decreases}, or \emph{impact}; unknown directions map to the zero vector), yielding a 1540-dimensional screen representation. This vector is projected to 256 dimensions by a linear layer followed by ReLU.

\paragraph{Gene representation.}
Each gene is represented by a stack of 43 pre-computed 128-dimensional gene embeddings~\citep{littman2025gene}. Missing gene--channel pairs are filled with zero vectors. A learned 16-dimensional positional embedding is appended to each channel token, producing 43 vectors of dimension 144.

\paragraph{DeepSet architecture.}
The 43 gene tokens are passed independently through a shared per-token network $\phi$ ($144 \to 256 \to 128$, ReLU, dropout) and mean-pooled into a single 128-dimensional gene vector. This vector is concatenated with the 256-dimensional projected screen representation and fed through a prediction network $\rho$ with hidden layers $384 \to 512 \to 256 \to 1$, using batch normalization, ReLU, and dropout ($p{=}0.1$).

\paragraph{Training.}
The model is trained with mean squared error on the continuous relevance scores using AdamW (learning rate $10^{-3}$, weight decay $10^{-4}$, batch size 512, 10 epochs). A reduce-on-plateau scheduler monitors the validation loss (factor 0.5, patience 5). The best checkpoint is selected by validation \andcg\at{100}, not by validation loss.

\paragraph{Architectural variants.}
We explored eight configurations on the temporal split by varying three binary factors: (i)~regression versus three-class classification (negative, zero, positive, with thresholds at $\pm 0.2$), (ii)~adding a learnable 128-dimensional gene-specific embedding as a 50th token to the DeepSet input, and (iii)~reweighting examples so that each screen contributes equally to the loss regardless of its gene count. The regression variant without learnable gene embeddings and without equal-screen weighting achieved the best validation \andcg\at{100} and is the model reported in the main text.

\subsection{Ensemble strategies}
\label{app:ensemble}

Due to the distribution shift between dataset splits, it is difficult to train a model which outperforms zero-shot frontier LLMs. Even naive few-shot approaches cause regression in AnDCG scores compared to zero-shot performance. Hence, to improve upon these base LLMs, we consider an ensemble approach. 
Initially, we learned an ensemble program using an evolutionary based approach with OpenEvolve \cite{openevolve} using GLM-5 \cite{zeng2026glm} as the evolver. Here, we evolve a function with a dictionary of LLM gene predictions as input and an ensembled list of gene predictions as output. Further, we also provide information on the screen, such as its phenotype and description, and we further provide the LLM access to tools to analyze possible programs using the training set. 
We evolved this program using validation set performance, where the goal was to maximize AnDCG@100. Finally, we computed test set performance to understand generalization of the program. Generally speaking, we find the evolution process to be fairly effective, and the ensemble performance on the validation set can improve over Gemini 3 Pro by up to $\sim$25\%. However, this performance increase generalizes poorly to the test set, with almost no performance improvement.

By studying the best performing ensembles proposed in the LLM evolution strategy, we identified reciprocal rank fusion \cite{cormack2009reciprocal} as a promising ensemble approach. To maximize performance of this ensemble function, we employed Bayesian optimization. We optimized over 1) individual LLM weights and $K_{RRF}$, 2) agreement bonuses between LLMs, and 3) nearest-neighbor data from the training set, as shown in the following program. 

\begin{lstlisting}[style=promptstyle]
def ensemble_predictions(
    all_model_predictions: Dict[str, List[List[str]]],
    neighbor_data: Optional[List[Dict[str, Any]]] = None,
    top_k: int = 100,
) -> List[str]:
    gene_scores: Dict[str, float] = defaultdict(float)
    gene_model_count: Dict[str, int] = defaultdict(int)

    for model_name, runs in all_model_predictions.items():
        w = MODEL_WEIGHTS.get(model_name, 1.0)
        if w < 0.01:
            continue
        for run in runs:
            for rank, gene in enumerate(run):
                g = gene.strip().upper()
                if g:
                    gene_scores[g] += w / (K_RRF + rank + 1)

        genes_from_model = set()
        for run in runs:
            for gene in run[:top_k]:
                genes_from_model.add(gene.strip().upper())
        for g in genes_from_model:
            gene_model_count[g] += 1

    for g, count in gene_model_count.items():
        if count >= AGREEMENT_THRESHOLD:
            if count >= 3:
                gene_scores[g] *= AGREEMENT_BONUS_3
            elif count >= 2:
                gene_scores[g] *= AGREEMENT_BONUS_2

    if neighbor_data and N_NEIGHBORS > 0:
        relevant_neighbors = [
            nb for nb in neighbor_data[:N_NEIGHBORS]
            if nb["similarity"] >= SIM_THRESHOLD
        ]

        if len(relevant_neighbors) >= KNN_BOOST_MIN_NEIGHBORS:
            knn_gene_evidence: Dict[str, float] = defaultdict(float)
            for nb in relevant_neighbors:
                sim = nb["similarity"]
                gt_genes = nb.get("ground_truth_genes", [])
                rel_scores = nb.get("relevance_scores", [])
                for j, gene in enumerate(gt_genes[:top_k]):
                    g = gene.strip().upper()
                    relevance = rel_scores[j] if j < len(rel_scores) else 1.0
                    knn_gene_evidence[g] += sim * relevance

            for g, evidence in knn_gene_evidence.items():
                if g in gene_scores:
                    gene_scores[g] *= (1.0 + KNN_BOOST_FACTOR * evidence)

            if KNN_ADD_ENABLED and len(relevant_neighbors) >= KNN_ADD_MIN_NEIGHBORS:
                for g, evidence in knn_gene_evidence.items():
                    if g not in gene_scores and evidence >= KNN_ADD_MIN_EVIDENCE:
                        gene_scores[g] = KNN_ADD_WEIGHT * evidence

    return sorted(gene_scores, key=gene_scores.get, reverse=True)[:top_k]

\end{lstlisting}

\begin{figure}[h]
    \centering
    \includegraphics[width=0.6\linewidth]{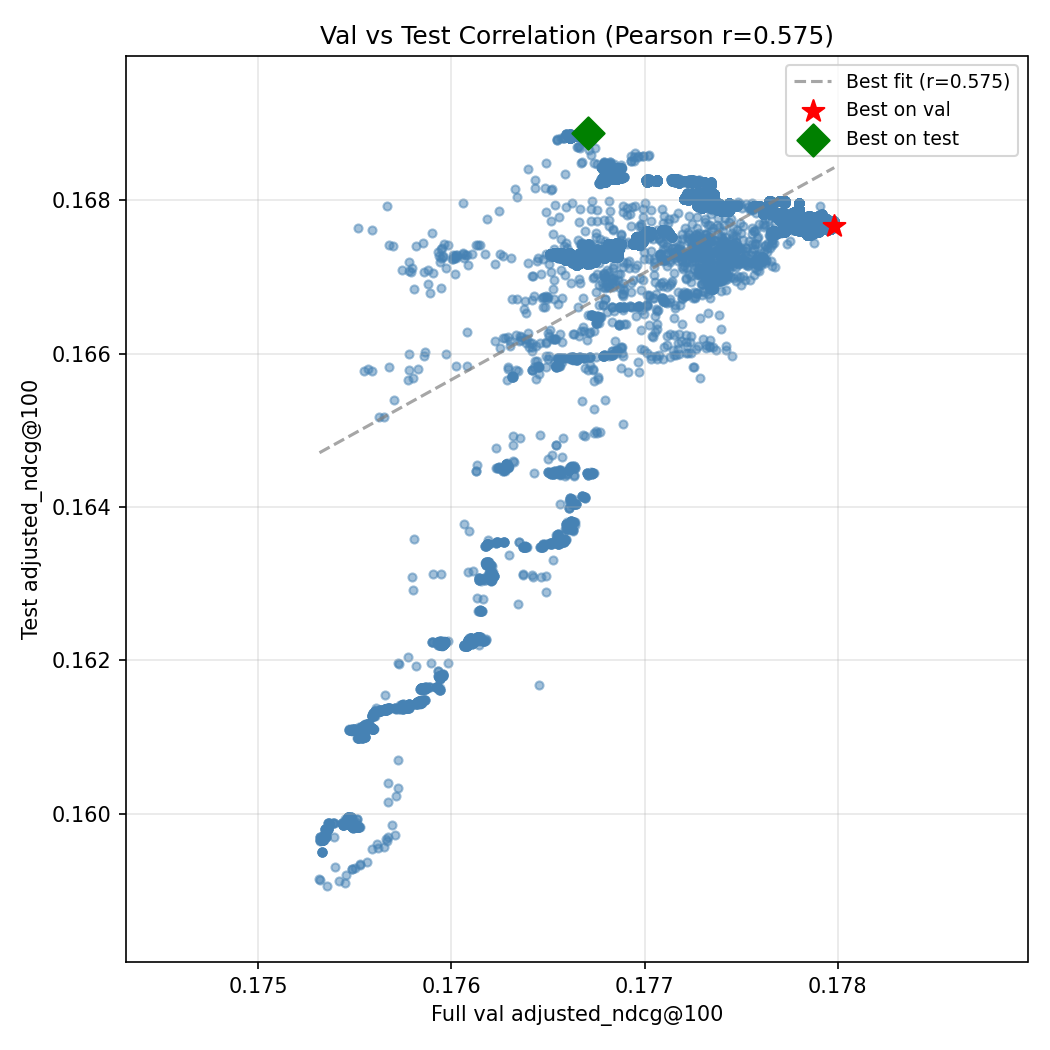}
    \caption{Validation vs. test performance of ensemble functions produced by Bayesian Optimization on validation AnDCG@100. The figure shows how optimizing validation performance results in suboptimal test performance. }
    \label{fig:bo_val_vs_test}
\end{figure}

Further analysis indicated that nearest-neighbor data and a large number of LLMs caused overfitting. Thus, we removed the nearest-neighbor data and restricted the model to three equally-weighted LLMs: gemini-3-flash, gemini-3-pro, and GPT-5.4. Finally, we observed reliable generalization to the test set, as shown in Figure \ref{fig:bo_val_vs_test}.

The LLM ensemble used in the main paper was selected as the function with best performance on the validation set, as shown in the following function. Overall, we found an ensemble method to be the most effective approach for outperforming zero-shot LLMs. However, future improvements will need to carefully consider solutions to the overfitting problem. 

\newpage

\begin{lstlisting}[style=promptstyle]
MODEL_WEIGHTS = {
"gemini-3-pro": 1.000000,
"gpt-5.4": 1.000000,
"gemini-3-flash": 1.000000,
}

K_RRF = 6.807906
AGREEMENT_BONUS_2 = 1.181246
AGREEMENT_BONUS_3 = 1.290464
AGREEMENT_THRESHOLD = 5

def ensemble_predictions(
    all_model_predictions: Dict[str, List[List[str]]],
    top_k: int = 100,
) -> List[str]:
    gene_scores: Dict[str, float] = defaultdict(float)
    gene_model_count: Dict[str, int] = defaultdict(int)

    for model_name, runs in all_model_predictions.items():
        w = MODEL_WEIGHTS.get(model_name, 1.0)
        if w < 0.01:
            continue
        for run in runs:
            for rank, gene in enumerate(run):
                g = gene.strip().upper()
                if g:
                    gene_scores[g] += w / (K_RRF + rank + 1)

        genes_from_model = set()
        for run in runs:
            for gene in run[:top_k]:
                genes_from_model.add(gene.strip().upper())
        for g in genes_from_model:
            gene_model_count[g] += 1

    for g, count in gene_model_count.items():
        if count >= AGREEMENT_THRESHOLD:
            if count >= 3:
                gene_scores[g] *= AGREEMENT_BONUS_3
            elif count >= 2:
                gene_scores[g] *= AGREEMENT_BONUS_2

    return sorted(gene_scores, key=gene_scores.get, reverse=True)[:top_k]

\end{lstlisting}

\subsection{Compute resources}

Proprietary language models were evaluating via their respective API. Open source languages models and neural gene-relevance predictors where evaluated on 16 NVIDIA B200 GPUs.

%% file: sections/appendix/D_results.tex
\section{Additional Results}
\label{app:additional_results}

\subsection{Performance Gains over Time}\label{sec:gains_over_time}

\begin{figure}[h]
    \centering
    \includegraphics[width=0.8\linewidth]{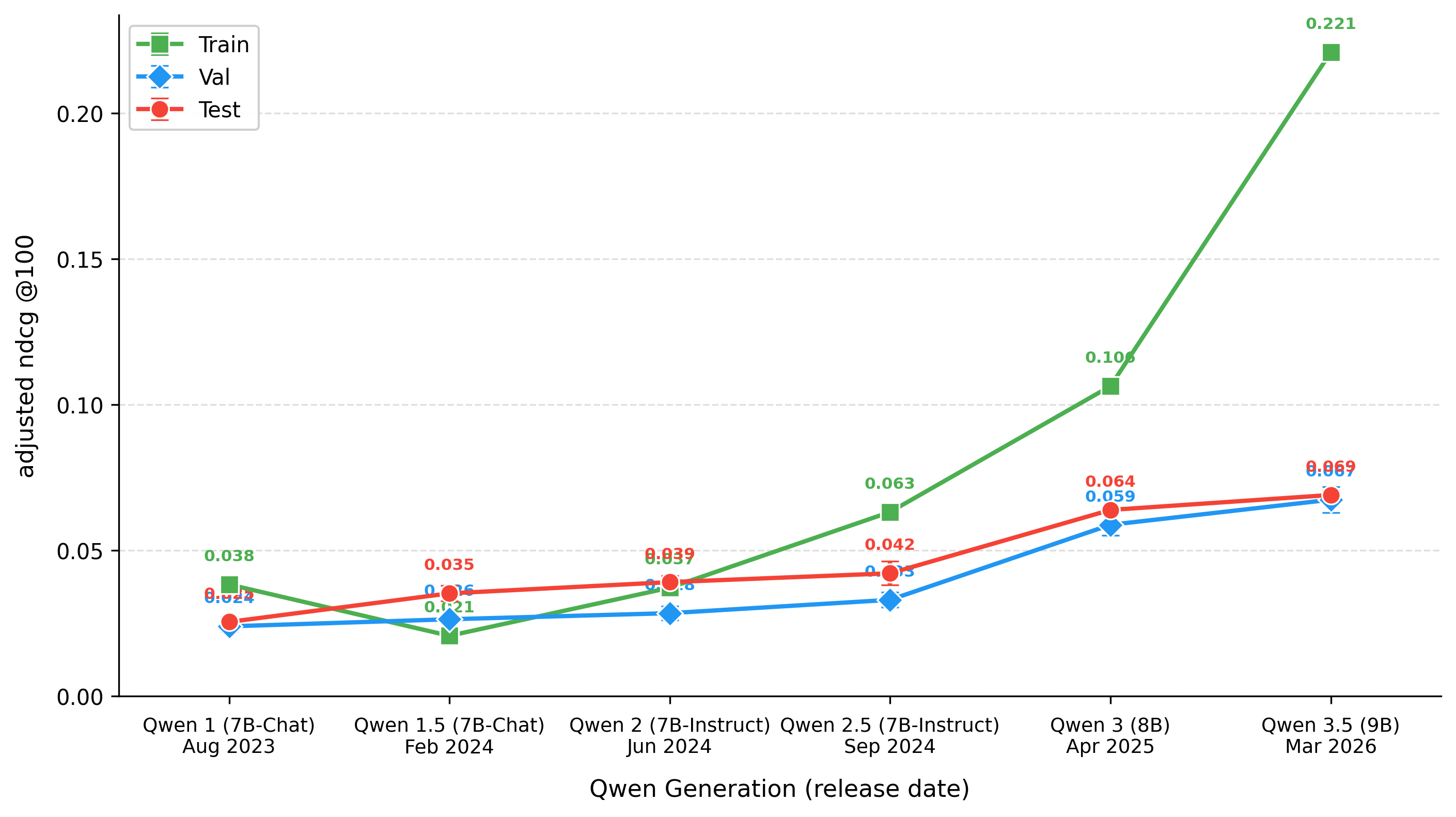}
    \caption{Qwen $\sim8$B model performance over time. }
    \label{fig:qwen_over_time}
\end{figure}

In addition to the parameter scaling results shown in Figure \ref{fig:figure4}, we also consider performance improvements over model generations. Due to API retirements, this is only possible with open models, so we select the Qwen family of models. We select $\sim8$B models since that size is the largest which was released across all generations. The overall trend (Figure \ref{fig:qwen_over_time}) indicates that improved datasets, architectures, and training procedures have increased the performance of similarly sized models. 
%Here, a key question to ask is whether this improvement is due to learning a better biological `world model', or simply memorization.
%\vspace{1em}

Here, a key question to ask is whether this improvement is due to learning a better biological `world model', or simply memorization.

%\vspace{1em}
Recent research \citep{li2026incompressible} suggests that rare facts are non-compressible, but procedural capability is compressible (following densing laws). Hence, the observed performance increase may indicate that the models are indeed learning to \textit{understand} biological effects better, rather than just memorizing non-compressible facts about screens. Alternatively, the scope of AssayBench (BioGRID ORCS) may just fall within the memorization capabilities of the model or reflect improved training data coverage.

\subsection{Per-Cohort Results for All Benchmarks}

Table~\ref{tab:year_metrics} reports per-cohort results for all benchmark models.

\input{tables/year_split_results}

%% file: tables/year_split_results.tex
% Requires \usepackage{booktabs}
\begin{table}[h!]
\centering
\small
\begin{tabular}{llrrr}
\toprule
Method & Cohort & \andcg\at{100} & \Prec\at{100} & \fdr\at{100} \\
\midrule
Gemini 3 Pro & val & 0.1716 & 0.3710 & 0.0208 \\
Gemini 3 Pro & test & 0.1570 & 0.2226 & 0.0164 \\
Gemini 3 Pro & \novel{} & 0.1113 & 0.2340 & NA \\
\midrule
GPT-5.4 & val & 0.1658 & 0.3262 & 0.0162 \\
GPT-5.4 & test & 0.1470 & 0.1980 & 0.0206 \\
GPT-5.4 & \novel{} & 0.0982 & 0.2218 & NA \\
\midrule
GPT-OSS-120B (SFT + GRPO) & val & 0.1361 & 0.2307 & 0.0165 \\
GPT-OSS-120B (SFT + GRPO) & test & 0.1293 & 0.1699 & 0.0252 \\
GPT-OSS-120B (SFT + GRPO) & \novel{} & 0.1012 & 0.2525 & NA \\
\midrule
Gene-frequency (by phenotype) & val & 0.1691 & 0.4103 & 0.0606 \\
Gene-frequency (by phenotype) & test & 0.1334 & 0.2204 & 0.0292 \\
Gene-frequency (by phenotype) & \novel{} & 0.0888 & 0.1866 & NA \\
\midrule
GPT-OSS-120B (SFT) & val & 0.1296 & 0.2273 & 0.0197 \\
GPT-OSS-120B (SFT) & test & 0.1224 & 0.1738 & 0.0255 \\
GPT-OSS-120B (SFT) & \novel{} & 0.1088 & 0.1921 & NA \\
\midrule
Biomni A1 (Claude 4) & val & 0.1410 & 0.2474 & 0.0164 \\
Biomni A1 (Claude 4) & test & 0.1079 & 0.1701 & 0.0176 \\
Biomni A1 (Claude 4) & \novel{} & 0.0728 & 0.0214 & NA \\
\midrule
Gene-relevance predictor & val & 0.1409 & 0.2155 & 0.0316 \\
Gene-relevance predictor & test & 0.0565 & 0.1577 & 0.0306 \\
Gene-relevance predictor & \novel{} & 0.0660 & 0.1541 & NA \\
\midrule
GPT-OSS-120B & val & 0.1268 & 0.2292 & 0.0164 \\
GPT-OSS-120B & test & 0.1211 & 0.1757 & 0.0223 \\
GPT-OSS-120B & \novel{} & 0.0826 & 0.2497 & NA \\
\midrule
C2S (Gemma-2B) & val & 0.0842 & 0.1200 & 0.0429 \\
C2S (Gemma-2B) & test & 0.0355 & 0.0863 & 0.0218 \\
C2S (Gemma-2B) & \novel{} & 0.0078 & 0.0784 & NA \\
\midrule
Oracle kNN & val & 0.3992 & 0.4755 & 0.0025 \\
Oracle kNN & test & 0.2918 & 0.3798 & 0.0037 \\
Oracle kNN & \novel{} & 0.2307 & 0.3079 & NA \\
\midrule
Embedding kNN & val & 0.0697 & 0.1027 & 0.0321 \\
Embedding kNN & test & 0.0646 & 0.1062 & 0.0226 \\
Embedding kNN & \novel{} & 0.0271 & 0.1127 & NA \\
\midrule
Gemini 3 Flash (GEPA) & val & 0.1922 & 0.3444 & 0.0102 \\
Gemini 3 Flash (GEPA) & test & 0.1406 & 0.1987 & 0.0214 \\
Gemini 3 Flash (GEPA) & \novel{} & 0.1397 & 0.3070 & NA \\
\midrule
Gemini 3 Flash & val & 0.1556 & 0.2495 & 0.0260 \\
Gemini 3 Flash & test & 0.1446 & 0.2009 & 0.0180 \\
Gemini 3 Flash & \novel{} & 0.0908 & 0.2351 & NA \\
\midrule
Gemini 3 Pro (Few-shot) & val & 0.1687 & 0.2876 & 0.0155 \\
Gemini 3 Pro (Few-shot) & test & 0.1537 & 0.2134 & 0.0213 \\
Gemini 3 Pro (Few-shot) & \novel{} & 0.1297 & 0.2449 & NA \\
\midrule
LLM Ensemble & val & 0.1864 & 0.3235 & 0.0145 \\
LLM Ensemble & test & 0.1631 & 0.2196 & 0.0178 \\
LLM Ensemble & \novel{} & 0.1088 & 0.2477 & NA \\
\midrule
Qwen3.5-2B & val & 0.0237 & 0.0695 & 0.0216 \\
Qwen3.5-2B & test & 0.0284 & 0.0755 & 0.0216 \\
Qwen3.5-2B & \novel{} & 0.0324 & 0.1342 & NA \\
\bottomrule
\end{tabular}
\caption{All benchmark models on the temporal split, reporting \andcg\at{100}, \Prec\at{100}, and \fdr\at{100} per cohort. \fdr{} is not defined for splits where no screen has negative relevance scores (NA).}
\label{tab:year_metrics}
\end{table}